\documentclass[acmsmall,screen]{acmart}
\AtBeginDocument{%
  }

\setcopyright{none}
\copyrightyear{2018}
\acmYear{2018}
\acmDOI{XXXXXXX.XXXXXXX}

\acmConference[Conference acronym 'XX]{Make sure to enter the correct
  conference title from your rights confirmation email}{June 03--05,
  2018}{Woodstock, NY}
\acmPrice{15.00}
\acmISBN{978-1-4503-XXXX-X/18/06}

\usepackage{amsmath,amsthm,amsfonts}

\usepackage{algorithm}
\usepackage{algorithmic}

\usepackage{listings}
\lstdefinestyle{jav}{language=Java,basicstyle=\ttfamily\footnotesize,mathescape,columns=flexible,keepspaces=true,numbers=left,numberstyle=\tiny,stepnumber=1,numbersep=5pt,escapechar=\%,xleftmargin=\leftmargini}
\lstnewenvironment{jav}
  {\lstset{style=jav}}
  {}

\usepackage{subcaption}
\usepackage{soul}
\usepackage{mathtools}

\usepackage{cleveref}
\crefname{theorem}{Thm.}{Thms.}
\crefname{lemma}{Lem.}{Lemmas}
\crefname{corollary}{Cor.}{Cors.}
\crefname{figure}{Fig.}{Figs.}
\crefname{definition}{Defn.}{Defns.}
\crefname{table}{Tab.}{Tabs.}
\crefformat{section}{\S#2#1#3}
\crefmultiformat{section}{\S#2#1#3}{ and~\S#2#1#3}{, \S#2#1#3}{ and~\S#2#1#3}
\crefname{example}{Ex.}{Exs.}
\crefname{item}{item}{items}
\crefname{footnote}{footnote}{footnotes}
\crefname{observation}{Obs.}{Obs.}
\crefname{remark}{Remark}{Remarks}
\crefname{proposition}{Prop.}{Props.}
\crefname{equation}{Eqn.}{Eqns.}
\crefname{counterexample}{Counterexample}{Counterexamples}
\crefname{property}{Property}{Properties}
\crefname{algorithm}{Algorithm}{Algorithms}
\usepackage{shortcuts}

\newcommand{\m}[1]{\mathsf{#1}}
\newcommand{\mi}[1]{\mathit{#1}}

\newcommand{\framework}{\textsc{DSE-SMC}}
\newcommand{\slowfuzz}{\textsc{SlowFuzz}}
\newcommand{\mayhem}{\textsc{Mayhem}}
\newcommand{\kelinci}{\textsc{Kelinci}}
\newcommand{\kelinciWCA}{\textsc{KelinciWCA}}
\newcommand{\badger}{\textsc{Badger}}
\newcommand{\libfuzzer}{\textsc{LibFuzzer}}
\newcommand{\evosuite}{\textsc{EvoSuite}}
\newcommand{\murphy}{\textsc{Murphy}}

\newcommand{\afl}{\textsc{AFL}}
\newcommand{\jdk}{\textsc{OpenJDK}}

\begin{document}

\title{Worst-Case Analysis is Maximum-A-Posteriori Estimation}
\subtitle{Resource Analysis with Sequential-Monte-Carlo-Based Fuzzing}

\author{Hongjun Wu}
\affiliation{%
 \institution{Peking University}
 \country{China}}

\author{Di Wang}
\affiliation{%
 \institution{Peking University}
 \country{China}}


\begin{abstract}
% State the problem
The worst-case resource usage of a program can provide useful
information for many software-engineering tasks, such as performance
optimization and algorithmic-complexity-vulnerability discovery.
This paper presents a generic, adaptive, and sound fuzzing
framework, called \framework{}, for estimating worst-case resource
usage.
%
% Say why it's an interesting problem
\framework{} is \emph{generic} because it is black-box as long as
the user provides an interface for retrieving resource-usage
information on a given input;
\emph{adaptive} because it automatically balances between
exploration and exploitation of candidate inputs;
and \emph{sound} because it is guaranteed to converge to the true
resource-usage distribution of the analyzed program.

% Say what your solution achieves
\framework{} is built upon a key observation: resource accumulation
in a program is isomorphic to the \emph{soft-conditioning} mechanism
in Bayesian probabilistic programming; thus, worst-case resource
analysis is isomorphic to the maximum-a-posteriori-estimation problem
of Bayesian statistics.
\framework{} incorporates sequential Monte Carlo (SMC)---a generic
framework for Bayesian inference---with adaptive evolutionary fuzzing
algorithms, in a sound manner, i.e., \framework{} asymptotically
converges to the posterior distribution induced by resource-usage
behavior of the analyzed program.
%
% Say what follows from your solution
Experimental evaluation on Java applications demonstrates that
\framework{} is significantly more effective than existing black-box
fuzzing methods for worst-case analysis.
\end{abstract}



%\keywords{real-time verification, sequential monte carlo, fuzzing, genetic}


\maketitle

\section{Introduction}
\label{Se:Intro}

% Three useful guides to the organization of the intro section
%
% Organization A (Simon Peyton-Jones)
%  o Describe the problem
%  o State your contributions
%  o Organization of the paper [if not able to cover point by point
%    in contributions]
% 
% Organization B (Nickolai Zeldovich)
%  o Elevator story -- high-level statement of the problem
%  o The problem in more technical terms
%  o Conventional wisdom: sketch of previous techniques and their
%    shortcomings
%  o Describe the solution to the problem as a black box,
%    so that it is clear what our solution offers over previous work
%  o Technical challenges to obtaining a solution (e.g., 1 paragraph
%    for each)
%  o State how we solve the challenges (at most a few paragraphs)
%  o Experimental highlights
%  o Contributions
%  o Organization of the paper 
%
% Organization C (Derek Dreyer)
%  o Context: Set the stage, motivate the general topic
%  o Gap: Explain your specific problem and why existing work does not
%    adequately solve it
%  o Innovation: State what you've done that is new, and explain how it
%    helps fill the gap

Software performance testing has always been a significant concern for software developers and testers. 
With the development of software science, the increase in software complexity drives people to seek for automatic methods. 
From an industrial point of view, it is routinely requested that units of an algorithms performance should be tested for evaluation and reconstruction. 
Yet, the process has been from bottom to top for so long that engineers failed to transform it into an automated process. 
On the other hand, theoretically, it is crucial to notice that, effective as methods based on machine learning now seen, the outcome may be unstable and unreliable in quality. 

In common sense, a stable algorithm is one that manages to control its worst-case resource consumption. 
Generally, developers and testers now utilize fuzzing methods for detecting software vulnerabilities.
To achieve better performances, the algorithm should be reliable, stable, and risk-resilient, bringing us to the need for testing. 
Therefore, it is essential to provide a theory and an applicable approach for generating the worst-case inputs of an algorithm that reveals its potential resource-usage profile objectively, rigorously, and meaningfully. 

So far, the number of frameworks in generating the worst-case inputs is relatively limited, compared to fuzzing frameworks.
On the one hand, there have been black-box fuzzing-based worst-case-analysis tools such as \slowfuzz{}~\cite{CCS:PZK17}.
Those approaches are generic in the sense that they do not require domain knowledge, but they might be ineffective
when the structure of the analyzed program becomes complex.
On the other hand, there have been white- and grey-box worst-case-analysis tools, many of which rely on symbolic execution~\cite{ICSE:BJS09,ICST:LKP17,POPL:WH19} or a combination of fuzzing and symbolic execution~\cite{ISSTA:NKP18}.
Those approaches turn out to be more effective because they can use information of
the concrete implementation of the analyzed program to guide their exploration in the space
of candidate worst-case inputs, but they also demand more computational resources to do program analysis, execution-path search, etc.

In this paper, we consider the problem of \emph{black-box} worst-case analysis.
The analysis should be \emph{generic}, in the sense that it does not require domain- or application-specific knowledge;
\emph{adaptive}, in the sense that it automatically balances between exploration and exploitation of the space of
candidate inputs; and \emph{sound}, in the sense that it correctly accounts for the resource-usage distribution of the analyzed program.
The analysis takes a resource-accumulating program (i.e., the program responds with its resource usage under a given metric) and a specification for candidate inputs, generates candidate inputs sequentially and learns from the information revealed in each generation to reach the worst-case resource behavior, and finally outputs an input with as large resource usage as possible. 

The major challenge is to actually learn something from the candidate inputs during the generation process,
given that the analysis is black-box.
Our key observation to solve the challenge is that as our title indicates, there is a correspondence between
\textbf{worst-case analysis (WCA) on resource-accumulating programs}
and
\textbf{maximum-a-posteriori (MAP) estimation on probabilistic programs}.
MAP estimation is a well-studied problem in Bayesian inference: it aims to find optimal parameters that maximize the posterior distribution
of a probabilistic model, conditioned on observations for the model.
Probabilistic programming~\cite{book:FoundProbProg20} provides a systematic way
of specifying probabilistic models as probabilistic programs, where special
\emph{soft-conditioning} statements account for observations and likelihood accumulation.
We thus develop an isomorphism between resource accumulation in an ordinary program
and likelihood accumulation in a probabilistic program.
Such a correspondence allows us to adapt advances in the field of Bayesian inference,
especially MAP estimation, to carry out worst-case analysis.

In this paper, we consider \emph{sequential Monte Carlo} (SMC), a versatile framework
for approximating and optimizing posterior distributions~\cite{JRSS:DDJ06}.
SMC works by maintaining a set of weighted samples and iteratively evolving them based on
observations for the probabilistic model.
The workflow of SMC is very similar to evolutionary fuzzing techniques, which
maintain a population of candidates and iteratively evolving them via genetic operations
such as crossover and mutation.
Therefore, we develop \underline{D}ual-\underline{S}trategy \underline{E}volutionary Sequential Monte-Carlo,
abbreviated by \framework{},
which incorporates SMC with evolutionary fuzzing.
The innovation of \framework{} lies in two main aspects.
Firstly, we integrate resample-move SMC~\cite{JRSS:GB01}, which allows using
arbitrary Markov-Chain Monte Carlo (MCMC) kernels to evolve the set of samples,
with evolutionary MCMC~\cite{EA:DT03}, which recasts genetic operations such as crossover
and mutation as Monte Carlo methods, and obtain a generic, adaptive, and sound
evolutionary SMC framework.
Secondly, we introduce a nature-inspired structure of simulating reproduction strategies of organisms~\cite{kn:AH86}
to strike a balance between exploration and exploitation.
The high-level idea is that organisms take different growth strategies in
crowded and uncrowded environments: in the former case, they prefer generating
a few offspring but with high quality, whereas in the latter case, they prefer
generating a lot of offspring mainly for increasing diversity.
We implement this idea as a dual-strategy approach, in the sense that
\framework{} maintains two groups of population, allows them to migrate from each other,
and evolves inside each group using a group-specific strategy.

\paragraph{Contributions}
The paper's contributions include the following:
\begin{itemize}
  \item We establish a correspondence between worst-case analysis (WCA) and maximum-a-posteriori (MAP)
  by showing how to reduce the WCA problem on a resource-accumulating program to the MAP problem on a probabilistic program, and vice versa (\cref{Se:WCAisMAP}).
  
  \item We devise \framework{}, an SMC-based fuzzing framework for WCA (\cref{Se:SMCFuzzingForWCA,Se:Technical}).
  
  \item We implemented a prototype of \framework{} and evaluated it on eight subjects. Our evaluation shows that \framework{} is significantly more effective than prior black-box methods (\cref{Se:Evaluation}).
\end{itemize}
\cref{Se:Related} discusses related work. \cref{Se:Conclusion} concludes. \cref{Se:DataAvailability} gives a statement of data availability.
\section{Overview}
\label{Se:Overview}

\subsection{Problem Statement}
\label{Se:ProblemStatement}

In this paper, we analyze the worst-case resource usage of a given
application by automatically generating inputs---of some given size---that
trigger as large resource usage as possible.
We assume that the user of our tool provides an interface to collect
resource-usage information from multiple executions of the application,
with possibly different inputs.
In this way, the user can specify a custom \emph{resource metric} of
interest, e.g., the number of executed program statements, jumps (branches),
or method calls.
Apart from the input specification and the resource-usage information,
we require that our tool be \emph{black-box}, i.e., it does not have
any application-specific knowledge for worst-case analysis.
Note that in this paper, our goal is \emph{not} to estimate asymptotic
complexities; instead, we focus on maximizing a user-specified concrete
metric over inputs of a given size.

\subsection{Worst-Case Analysis is Maximum-A-Posteriori Estimation}
\label{Se:WCAisMAP}

To demonstrate our framework, we consider insertion sort as a running example.
For an ordinary implementation of insertion sort, it is well known that its
worst-case time complexity is $O(n^2)$ and its best-case complexity
is $O(n)$, where $n$ is the length of the array-to-be-sorted.
\cref{Fi:RunningExample}(a) presents an implementation of insertion sort
in Java.
To make the resource metric \emph{explicit}, we interface resource accumulation
with method calls of the form \verb|Resource.tick(int)|.
For the program in \cref{Fi:RunningExample}(a), the resource metric accounts for
the total number of loop iterations.
For example, if we want to analyze the worst-case resource usage when
the length of the input array is $5$, a worst-case input could be
$[5,4,3,2,1]$ with a resource usage of $14$ loop iterations.

\begin{figure}
\centering
\begin{subfigure}[b]{0.48\textwidth}
\begin{jav}
public static void sort(int[] a) {
  int n = a.length;
  for (int i = 1; i < n; i ++) {
    int j = i - 1;
    int x = a[i];
    while (j >= 0 && a[j] > x) {
      a[j + 1] = a[j];
      j --;
      %\hl{Resource.tick(1);}%
    }
    a[j + 1] = x;
    %\hl{Resource.tick(1);}%
  }
}
\end{jav}
\caption{An implementation in Java}
\end{subfigure}
\begin{subfigure}[b]{0.5\textwidth}
\begin{jav}
public static void sort(int[] a) {
  int n = a.length;
  for (int i = 1; i < n; i ++) {
    int j = i - 1;
    int x = a[i];
    while (j >= 0 && a[j] > x) {
      a[j + 1] = a[j];
      j --;
      %\hl{Probability.score(Math.exp(1));}%
    }
    a[j + 1] = x;
    %\hl{Probability.score(Math.exp(1));}%
  }
}
\end{jav}
\caption{Resource accumulation as soft conditioning}
\end{subfigure}
\caption{Running example: insertion sort}
\label{Fi:RunningExample}
\end{figure}

To show how the worst-case analysis (WCA) problem is \emph{isomorphic} to the
maximum-a-posteriori (MAP) estimation, we first review Bayesian inference
and probabilistic programming.
\begin{itemize}
  \item \textbf{Bayesian inference} is a method for inferring the posterior
  distribution of a probabilistic model conditioned on observed data, with
  applications in artificial intelligence~\cite{NATURE:Ghahramani15},
  cognitive science~\cite{kn:GKT08}, applied statistics~\cite{book:GCS13}, etc.
  At the core of Bayesian inference is Bayes' law:
  \begin{equation}\label{Eq:BayesLaw}
  P(X=x \mid Z=z) = \frac{ P(Z=z \mid X=x) \cdot  P(X=x) }{ P(Z=z) },
  \end{equation}
  where $X$ and $Z$ are random variables standing for parameters and
  observations, respectively; $P(X =x \mid Z = z)$ is the conditional
  probability of parameters being $x$ given that observations are $z$,
  i.e., the \emph{posterior}; $P(Z = z \mid X = x)$ is the conditional
  probability of observations being $z$ given that parameters are $x$,
  i.e., the \emph{likelihood}; and $P(X=x)$ is the probability of
  parameters being $x$, i.e., the \emph{prior}.
  Because Bayesian inference usually fixes the observations $z$,
  the denominator $P(Z=z)$ of the right-hand side of \cref{Eq:BayesLaw}
  is usually considered as a constant, and thus
  we can write \cref{Eq:BayesLaw} as
  \begin{equation}\label{Eq:BayesLawProp}
  P(X = x \mid Z = z) \propto P(Z = z \mid X = x) \cdot P(X = x) .
  \end{equation}
  Bayesian inference usually accounts for the posterior distribution
  $P(X=x \mid Z=z)$ as a function of $x$ given some $z$.
  People have developed many algorithms to sample from the posterior
  distribution, e.g., Markov-Chain Monte Carlo (MCMC).
  In some other scenarios, people have also considered the
  \emph{maximum-a-posteriori} (MAP) estimation, i.e., finding an $x^*$
  that maximizes the posterior probability $P(X=x^* \mid Z =z)$.
  
  \item \textbf{Probabilistic programming} provides a flexible way of
  specifying probabilistic models and performing Bayesian inference~\cite{book:FoundProbProg20}.
  One way to understand the semantics of a probabilistic program is
  that it describes the measure that is induced by the product of the
  likelihood and the prior, i.e., the right-hand side of \cref{Eq:BayesLawProp}.
  Note that because we ignore the denominator $P(Z=z)$, the product
  is usually \emph{not} a probability measure.
  Consequently, probabilistic programming languages (PPLs)---such as
  Stan~\cite{JSS:CGH17}, Pyro~\cite{JMLR:BCJ18}, Church~\cite{UAI:GMR08,misc:dippl},
  and Gen~\cite{PLDI:CSL19}---have devised many techniques to
  sample from or approximate unnormalized measures defined by probabilistic
  programs.
  A PPL is usually an ordinary programming language with two special
  extra constructs:
  \begin{itemize}
    \item \emph{sampling}, which draws a random value from a prior distribution; and
    \item \emph{soft conditioning}, which records the likelihood of some observation.
  \end{itemize}
  Below gives a simple probabilistic program written as Java pseudocode.
  The program models a Bayesian-inference task: (i) there is an imprecise scale
  that responds with a noisy measurement from the Normal distribution
  $\calN(w,0.75^2)$ when the actual weight is $w$; (ii) we use the scale to
  weigh an object and read $9.5$ as the measurement; (iii) we have a prior
  guess that the object's weight is around $8.5$, modeled with the prior
  distribution $\calN(8.5,1^2)$; and (iv) we want to know the posterior
  distribution of the object's weight.
\begin{jav}
float $w$ = Probability.sample(Distribution.Normal($8.5$, $1$));
Probability.observe($9.5$, Distribution.Normal($w$, $0.75$));
\end{jav}
  The statement $\m{observe}(\mi{valu}, \mi{dist})$ is usally a wrapper around
  the more primitive statement $\m{score}(\ell)$, where $\ell$ is the likelihood
  of $\mi{valu}$ being sample from $\mi{dist}$; in this example, we can compute
  the likelihood using the probability density function of Normal distributions:
\begin{jav}
float $w$ = Probability.sample(Distribution.Normal($8.5$, $1$));
Probability.score( $\frac{1}{\sqrt{2 \pi \cdot 0.75^2}} e^{ -\frac{(9.5-w)^2}{2 \cdot 0.75^2} }$ );
\end{jav}
  A probabilistic program can contain multiple soft-conditioning/scoring statements;
  intuitively, the likelihood of an execution path is the product of scores
  along the path.
\end{itemize}

Readers might already notice the resemblance between resource
accumulation (via \verb|Resource.tick|) and likelihood accumulation
(via \verb|Probability.score|).
The major difference here is that the resource usage of an execution path
is the \emph{sum} of ticks along the path, whereas the likelihood of an
execution path is the \emph{product} of scores along the path.
Observing that likelihoods from probabilistic programming are always \emph{positive},\footnote{%
Some PPLs use zero likelihoods to enforce \emph{hard} constraints. In this paper, we only consider \emph{soft} constraints and thus we can assume that all likelihoods are non-zero.}
we derive a correspondence between WCA and MAP as follows.
\begin{itemize}
  \item \textbf{From WCA to MAP.} Given a program $M$ and a resource metric
  $\mi{tick} : \Sigma \to \bbR$ that assigns resource usages to program instructions
  in $\Sigma$,  the WCA problem aims to find an input $\theta$ for $M$ such
  that $\mi{tick}_M(\theta) \defeq \sum_{\sigma \in \pi(M,\theta)} \mi{tick}(\sigma)$ is maximized,
  where $\pi(M,\theta)$ gives the execution path of $M$ with $\theta$ as its input.
  Define a likelihood assignment $\mi{score} : \Sigma \to \bbR_{>0}$ as
  $\mi{score}(\sigma) \defeq e^{\mi{tick}(\sigma)}$ for any $\sigma \in \Sigma$.
  The map $\mi{score}_M \defeq \theta \mapsto \prod_{\sigma \in \pi(M,\theta)} \mi{score}(\sigma)$
  then defines a probabilistic semantics with no prior, where the MAP problem
  aims to find an input $\theta$ for $M$ such that $\mi{score}_M(\theta)$ is maximized.
  Because $\mi{score}_M(\theta) = \prod_{\sigma \in \pi(M,\theta)} \mi{score}(\sigma) = \prod_{\sigma \in \pi(M,\theta)} e^{\mi{tick}(\sigma)} = e^{\sum_{\sigma \in \pi(M,\theta)} \mi{tick}(\sigma)} = e^{\mi{tick}_M(\theta)}$ for any $\theta$,
  a solution to the MAP problem is also a solution to the WCA problem.
  
  \item \textbf{From MAP to WCA.} Given a program $M$ and a likelihood assignment
  $\mi{score} : \Sigma \to \bbR_{>0}$ where $\Sigma$ is the set of program
  instructions.
  Conceptually, we can model probabilistic sampling with a pre-specified \emph{trace},
  in the sense that each sampling statement reads a value from the trace and
  records the prior probability accordingly~\cite{ICFP:BLG16,JCSS:Kozen81}.
  Thus, we can treat the trace as an input and the prior as a map $\mi{prior}$ such that
  $\mi{prior}(\tau)$ gives the prior probability of the trace $\tau$.
  The MAP problem then aims to find a trace $\tau$ such that $\mi{score}_M(\tau) \defeq \mi{prior}(\tau) \cdot \prod_{\sigma \in \pi(M,\tau)} \mi{score}(\sigma)$ is maximized, where---similarly to the former case---$\pi(M,\tau)$
  gives the execution path of $M$ with $\tau$ as its trace.
  As discussed above, the probabilities are positive, so we can define a resource
  metric $\mi{tick} : \Sigma \to \bbR$ as $\mi{tick}(\sigma) \defeq \log \mi{score}(\sigma)$ for
  any $\sigma \in \Sigma$.
  The map $\mi{tick}_M \defeq \tau \mapsto \log \mi{prior}(\tau) + \sum_{\sigma \in \pi(M,\tau)} \mi{tick}(\sigma)$
  then defines the resource accumulation of $M$ under the metric $\mi{tick}$,
  with the understanding that a resource usage of $\log \mi{prior}(\tau)$ is triggered at the
  beginning of a program execution.
  In this case, the WCA problem aims to find a trace $\tau$ for $M$ such that $\mi{tick}_M(\tau)$
  is maximized.
  Because $\mi{tick}_M(\tau) = \log \mi{prior}(\tau) + \sum_{\sigma \in \pi(M,\tau)} \mi{tick}(\sigma)
  = \log \mi{prior}(\tau) + \sum_{\sigma \in \pi(M,\tau)} \log \mi{score}(\sigma)
  = \log \left( \mi{prior}(\tau) \cdot \prod_{\sigma \in \pi(M,\tau)} \mi{score}(\sigma) \right)
  = \log \mi{score}_M(\tau)$ for any $\tau$, a solution to the WCA problem is also a solution to
  the MAP problem.
\end{itemize}
\cref{Fi:RunningExample} shows a direct demonstration of the correspondence.
In \cref{Fi:RunningExample}(a), we use \verb|Resource.tick(1)| to accumulate resource usages,
and in \cref{Fi:RunningExample}(b), we use \verb|Probability.score(Math.exp(1))| to accumulate likelihoods.
An MAP estimation for the probabilistic program in \cref{Fi:RunningExample}(b) when the length
of the input array is $5$ could be $[5,4,3,2,1]$, which is indeed a worst-case input for the
program in \cref{Fi:RunningExample}(a)---as we discussed at the beginning of \cref{Se:WCAisMAP}.

\subsection{Sequential-Monte-Carlo-Based Fuzzing for WCA}
\label{Se:SMCFuzzingForWCA}

Because worst-case analysis (WCA) is maximum-a-posteriori (MAP) estimation,
our goal is then to adapt MAP algorithms from Bayesian inference to carrying out WCA.
In this paper, we incorporate sequential Monte Carlo (SMC), evolutionary
algorithms, and fuzzing techniques to develop our \framework{} framework.
We start with a review of SMC and gradually extend it to present \framework{}.

\textbf{Sequential Monte Carlo} (SMC) methods form a
powerful family of Bayesian-inference algorithms for sampling from a \emph{sequence}
of target distributions~\cite{JRSS:DDJ06}.
Particle filters are a prominent SMC method for online inference in
state-space models, widely applied in tasks such as robot localization~\cite{book:ProbRobot05}.
An SMC method usually maintains a set of weighted samples as an empirical
approximation of the target distribution.
During each iteration of the sequential model, SMC uses a \emph{proposal}
distribution to generate new samples from previous ones and reweights the
new samples according to their likelihoods on the observed data.
When some samples have relatively negligible weights, SMC uses a resampling
process in which more promising samples are selected as the basis for future
inference.
Well-designed proposal distributions can bring significant performance
improvements~\cite{NIPS:GGT15, SYSID:WML18}.

Besides sequential modeling, when there is just a single target distribution
(e.g., the posterior distribution in the standard setting of Bayesian
inference), SMC has been shown to be a promising approach for sampling
from distributions with multiple modes~\cite{BS:DDJ07,Biometrika:Chopin02,ICML:SPH23,AWR:ZLM18}.
This property renders SMC desirable in our setting of worst-case analysis,
because worst-case inputs can usually be fairly separated among the input
space and long-tailed in the resource-usage distribution.
For example, any decreasing sequence of integers is a worst-case input
for the insertion-sort implementation in \cref{Fi:RunningExample}(a).

In this paper, we adapt a variant of SMC algorithms that is usually called
\emph{resample-move} SMC~\cite{JRSS:GB01}.
Consider a probabilistic semantics $\mi{score}_M : \bbS \to \bbR_{>0}$
for some probabilistic program $M$, where $\bbS$ is its input space.
The SMC algorithm is overall iterative: at the $t$th epoch, it maintains a set
$\{(\theta_t^\ell,w_t^\ell)\}_{\ell=1}^L$ of $L \ge 1$ weighted samples,
where the weight $w_t^\ell > 0$ reflects the likelihood of
the sample $\theta^\ell_t \in \bbS$, i.e., $\mi{score}_M(\theta^\ell_t)$,
relative to other samples.
Below outlines how the SMC algorithm proceeds.
The step (S0) initializes a set of $L$ samples randomly.
The $t$th epoch of the algorithm then involves three steps:
(S1) reweights each sample by its likelihood with respect to the probabilistic semantics,
(S2) resamples the samples based on their weights, and
(S3) rejuvenates the samples by running $n \ge 1$ iterations of
\textsc{GenerateNewSample} that implements a MCMC transition kernel.
\[
\begin{array}{|r|l|}
  \hline
  \text{(S0)} & \text{At epoch $t=0$, initialize $\theta^\ell_0$ randomly for each $\ell=1,\ldots,L$.} \\[8pt]
  & \text{For each epoch $t>0$, run steps (S1) to (S3):} \\[4pt]
  \text{(S1)} & \text{\textbf{Reweight:} For $\ell =1,\ldots,L$, compute likelihoods as the weights:} \\[4pt]
  & \qquad w_t^\ell \gets \mi{score}_M(\theta_{t-1}^\ell)  \\[4pt]
  \text{(S2)} & \text{\textbf{Resample:} For $\ell=1,\ldots,L$, resample the parents and reset the weights:} \\[4pt]
  & \qquad u \sim \mathrm{Categorical}(w_t^1, \ldots, w_t^L), \enskip \theta^\ell_t \gets \theta^u_{t-1} \\[4pt]
  \text{(S3)} & \text{\textbf{Rejuvenate:} For $\ell=1,\ldots,L$, rejuvenate $\theta^\ell_t$ by running MCMC:} \\[4pt]
  & \qquad \theta^\ell_t \sim \textsc{GenerateNewSample}(\theta^\ell_t), \enskip \text{for $n \ge 1$ iterations} \\[4pt]
  \hline
\end{array}
\]
The reweighting step (S1), when considered in a WCA setting, computes $e^{\mi{tick}_M(\theta)}$ as the
weight for $\theta$, i.e., this step intuitively prioritizes inputs with large resource usages.
In the resampling step (S2), people have been using an \emph{adaptive} resampling strategy,
where resampling is triggered at epoch $t$ if the effective sample size---a commonly used
metric for sample diversity---drops under a threshold:
$\mathrm{ESS}(w_t^{1:L}) \defeq 1 / \sum_{\ell=1}^L \left(\frac{w_t^\ell}{ \sum_{k=1}^L w_t^k }\right)^2$.
The rejuvenating step (S3) makes use of arbitrary MCMC kernels to update the samples
and thus renders the SMC algorithm \emph{generic} because the implementation of
\textsc{GenerateNewSample} can be application-specific.
The \emph{soundness} of the SMC algorithm arises from its convergence properties, i.e.,
when the number of samples $L$ approaches infinity, the empirical approximation
$\{(\theta_t^\ell,w_t^\ell)\}_{\ell=1}^L$ asymptotically converges to the target
distribution given by $\mi{score}_M$.
For example, if the normalizing constant for the measure given by $\mi{score}_M$
is $Z_M$, the SMC algorithm guarantees that $\frac{1}{L} \sum_{\ell=1}^L \frac{w_t^\ell}{\sum_{k=1}^L w_t^k} \phi(\theta_t^\ell) \xlongrightarrow{L \to \infty} \sum_{\theta \in \bbS} \frac{\mi{score}_M(\theta)}{Z_M} \phi(\theta)$ almost surely, for any $\phi : \bbS \to \bbR$ that satisfies a few conditions~\cite{JRSS:GB01,JRSS:DDJ06}.

So far we have reviewed SMC for approximating posterior distributions, but our goal
is to use SMC for MAP estimation, which is an \emph{optimization} problem.
We now consider integrating SMC with \textbf{Evolutionary Algorithms} (EA), a popular family
of optimization algorithms that take inspiration from the biological evolution process.
A powerful class of EA is genetic algorithms (GA)~\cite{book:Goldberg89},
which resemble the natural selection process, and proceed by maintaining a population
of candidate solutions and evolving them via genetic operations, such as
(i) \emph{selection}, which selects candidates from the population based on
their \emph{fitness}, i.e., how well they optimize the objective,
(ii) \emph{crossover}, which simulates reproduction by taking two selected
candidates and mixing them to create offspring,
and (iii) \emph{mutation}, which allows random changes in the candidate solutions
to maintain the diversity of the population.

It is worth noting that SMC algorithms and genetic algorithms
have many similarities~\cite{RMTA:DKR01}.
Recall the resample-move SMC algorithm we just reviewed:
it maintains a set of weighted samples, as GA maintains a population of
candidates;
its reweighting step computes the likelihoods for samples, as GA computes
the fitness scores for candidates;
its resampling step randomly picks samples with higher likelihoods, as GA's
selection prioritizes candidates with higher fitness scores;
and its rejuvenating step randomly generate new samples from existing ones,
as GA's crossover and mutation evolve the population.
There have been studies on the integration of SMC and GA (or EA)~\cite{IROS:KFZ05,AWR:ZLM18,Econometrics:Dufays16},
which show that such integration is a promising approach for both
posterior approximation and MAP estimation.

In this paper, we develop \framework{} upon the generic resample-move SMC
algorithm and integrate techniques from EA in the resampling and rejuvenating steps.
\framework{} incorporates two ideas:
\begin{itemize}
  \item \textbf{\underline{D}ual \underline{S}trategy.}
  Optimization algorithms usually need to take care of the balance between
  exploration and exploitation of the solution space.
  \framework{} uses a dual-strategy approach inspired from 
  $K$-strategy and $R$-strategy~\cite{kn:AH86}, which
  are usually referred to two poles that describe the growth and reproduction
  strategies of organisms.
  Intuitively, $K$-strategy and $R$-strategy correspond to
  crowded and uncrowded environments, respectively.
  In terms of optimization, $K$-strategy tries to produce
  a few offspring with high fitness scores, whereas $R$-strategy
  tries to produce a lot of offspring to increase diversity.
  \framework{} splits the weighted samples into multiple groups,
  each of which uses $K$-strategy or $R$-strategy for rejuvenation.
  In each epoch, \framework{} allows \emph{migration} among the
  groups.

  \item \textbf{\underline{E}volutionary MCMC.}
  \framework{} adapts evolutionary MCMC~\cite{EA:DT03} as an
  adaptive and generic approach for implementing rejuvenation,
  i.e., the \textsc{GenerateNewSample} routine.
  MCMC is a general framework to sample from unnormalized
  measures such as the one induced by $\mi{score}_M$.
  The overall idea is to construct a Markov chain with the target
  distribution as the chain's stationary distribution; thus,
  simulating the chain generates correctly-distributed samples.
  We consider the Metropolis-Hastings (MH) algorithms for MCMC:
  let $\mi{proposal}(\theta_\m{new} ; \theta_\m{old})$ be
  a proposal distribution that generates a candidate new sample
  from an old one, then the new sample is accepted with probability
  $\min(1, \frac{\mi{score}_M(\theta_\m{new}) \cdot \mi{proposal}(\theta_\m{old} ; \theta_\m{new} ) }{ \mi{score}_M(\theta_\m{old}) \cdot \mi{proposal}(\theta_\m{new}; \theta_\m{old}) })$.
  It has been shown that both mutation and crossover can be recasted
  in terms of MH~\cite{EA:DT03,SS:LW00,Biometrika:JSH07}.
  For mutation, the proposal distribution simply implements how the random
  changes are applied to existing candidates.
  For crossover, one can have a crossover proposal distribution
  $\mi{proposal}(\tuple{\theta_{\m{new},1}, \theta_{\m{new},2}}; \tuple{ \theta_{\m{old},1}, \theta_{\m{old},2} })$
  and then the acceptance ratio is computed as
  $\min(1, \frac{ \mi{score}_M(\theta_{\m{new}_1}) \cdot \mi{score}_M(\theta_{\m{new},2}) \cdot \mi{proposal}( \tuple{\theta_{\m{old},1}, \theta_{\m{old},2}} ; \tuple{ \theta_{\m{new},1}, \theta_{\m{new},2} } ) }{ \mi{score}_M(\theta_{\m{old}_1}) \cdot \mi{score}_M(\theta_{\m{old},2}) \cdot \mi{proposal}( \tuple{\theta_{\m{new},1}, \theta_{\m{new},2}} ; \tuple{ \theta_{\m{old},1}, \theta_{\m{old},2} } ) })$.
  It is also possible for the population size to change along
  epochs~\cite{EA:DT03}; thus, \framework{} allows the set of
  weighted samples to have a dynamic size.
\end{itemize}

We have shown the key components of \framework{}, a generic, adaptive,
and sound framework for MAP estimation.
The final step of our development is then instantiating the framework to
carry out worst-case analysis (WCA) of resource usage.
As discussed above, the implementation of the rejuvenation step can be
application-specific; thus, in the WCA setting, we apply \textbf{fuzzing techniques}~\cite{ACS:ZWC22}
to implement the \textsc{GenerateNewSample} routine.
In particular, we adapt crossover and mutation operations from evolutionary
fuzzers such as \afl{}~\cite{misc:AFL} and \libfuzzer{}~\cite{misc:LibFuzzer} to manipulate program inputs
of different types, e.g., strings, integers, and arrays.
In our implementation, we recast those genetic operations as MCMC kernels.

\section{Technical Details}
\label{Se:Technical}

\begin{figure}
    \centering
	\includegraphics[width=\textwidth]{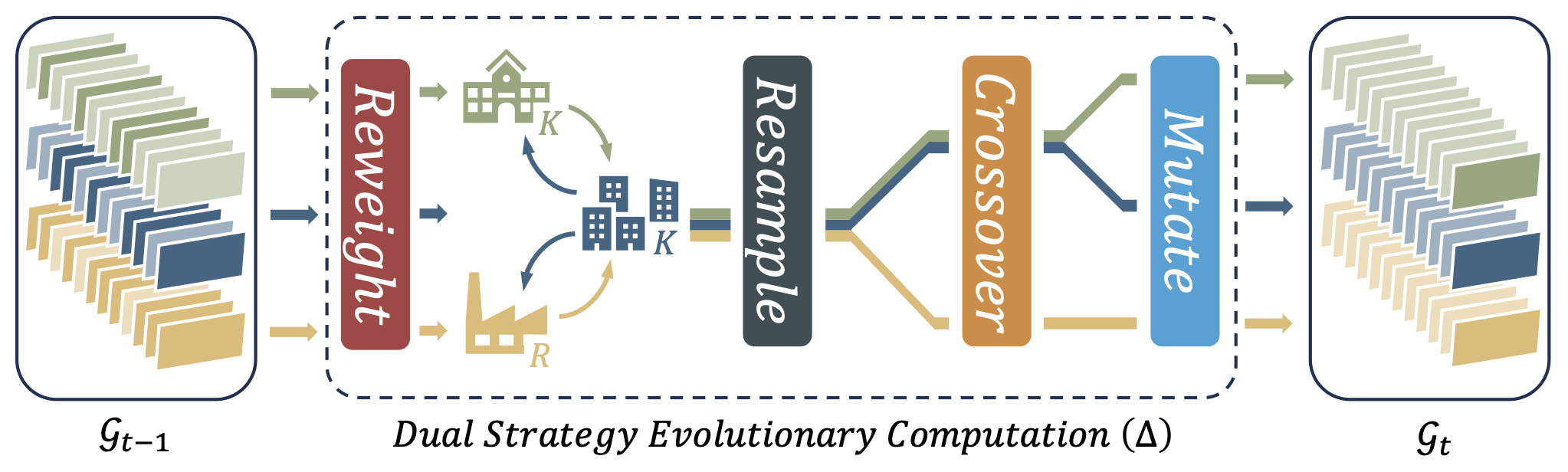}
	\caption{An epoch in \framework{}: reweight, migrate, resample, crossover, and mutate}
	\label{Fi:Workflow}
\end{figure}

\cref{Fi:Workflow} illustrates the workflow of an epoch in \framework{}.
As discussed in \cref{Se:SMCFuzzingForWCA}, \framework{} is an iterative framework:
at epoch $t$, it maintains a set $\calG_t$ of weighted samples;
and for $t>0$, it runs five steps to evolve from $\calG_{t-1}$ to $\calG_t$, namely
\emph{reweight}, \emph{migrate}, \emph{resample}, \emph{crossover}, and \emph{migrate}.
\cref{Alg:Framework} shows the pseudocode of our \framework{} framework with more
algorithmic details.
The algorithm takes a black-box probabilistic semantics $\mi{score}_M : \bbS \to \bbR_{>0}$
as its input, where $M$ stands for a probabilistic program and $\bbS$ is its input space.
In the setting of worst-case analysis, $M$ is usually an ordinary program with a
resource-accumulation semantics $\mi{tick}_M : \bbS \to \bbR$ and we simply define
$\mi{score}_M \defeq \theta \mapsto e^{\mi{tick}_M(\theta)}$.
Our approach is dual-strategy, meaning that we segregate candidate samples into
two groups, one for applying $K$-strategy, the other for $R$-strategy.
Different strategies mean that we apply different crossover and mutation operations, as
we will discuss later in this section.
Notationally, we write $\theta_t^\ell$ for the $\ell$th sample during the $t$th epoch,
as well as $\gamma_t^\ell \in \{K,R\}$ and $w_t^\ell>0$ for its group assignment and weight (or fitness score),
respectively.
We then write $\Theta_t$, $\Gamma_t$, and $W_t$ to be $\{ \theta_t^1,\ldots, \theta_t^{L_t} \}$,
$\{ \gamma_t^1,\ldots,\gamma_t^{L_t} \}$, and $\{ w_t^1, \ldots, w_t^{L_t} \}$, respectively,
where $L_t$ is the number of samples during the $t$th epoch.

We then discuss the five steps in an epoch of \framework{} one-by-one.

\begin{algorithm}
  \small
    \caption{\framework{}}
    \label{Alg:Framework}
    \begin{algorithmic}[1]
        \REQUIRE{A probabilistic semantics $\mi{score}_M : \bbS \to \bbR_{>0}$}
        \REQUIRE{Initial population size $L_0$ and population threshold $L_\m{max}$}
        \REQUIRE{An effect-sample-size threshold $\mathrm{ESS}_\m{min} : \bbN \to \bbN$ dependent on the number of samples}
        \STATE{$\rhd$ Initialize a population of random candidate samples with random group assignments}
        \STATE{$\Theta_0 \gets \{ \theta_{0}^1, \ldots, \theta_{0}^{L_0} \}$ where each $\theta_0^\ell \in \bbS$}
        \STATE{$\Gamma_0 \gets \{ \gamma_0^1,\ldots, \gamma_0^{L_0} \}$ where each $\gamma_0^\ell \in \{ K,R\}$}
        \STATE{$\rhd$ Main loop}
        \STATE{$t \gets 0$}
        \LOOP
        \STATE{$t \gets t + 1$}
        \STATE{$\rhd$ Reweight}
        \STATE{$W_{t-1} \gets \{ \mi{score}_M(\theta_{t-1}^\ell) \mid \ell = 1,\ldots,L_{t-1} \}$}
        \STATE{$\rhd$ Migrate}
        \STATE{$\Gamma_{t-1} \sim \textsc{Migrate}(W_{t-1},\Gamma_{t-1})$}
        \STATE{$\rhd$ Resample}
        \IF{$L_{t-1} > L_\m{max}$ or $\mathrm{ESS}(W_{t-1}) < \mathrm{ESS}_\m{min}(L_{t-1})$}
          \STATE{$L_t \gets \min(L_{t-1}, L_\m{max})$}
          \FOR{$\ell=1$ \TO $L_t$}
            \STATE{$u \sim \mathrm{Categorical}(W_{t-1})$}
            \STATE{$\theta_t^\ell, \gamma_t^\ell \gets \theta_{t-1}^u, \gamma_{t-1}^u$}
          \ENDFOR
        \ELSE
          \STATE{$\Theta_t,\Gamma_t \gets \Theta_{t-1},\Gamma_{t-1}$}
        \ENDIF
        \STATE{$\rhd$ Rejuvenate in each group}
        \FOR{$g \in \{K,R\}$}
          \STATE{$\tilde{\Theta}[g] \gets \{ \theta_t^\ell \mid \gamma_t^\ell = g \}$}
          \STATE{$\tilde{\Theta}[g] \sim \textsc{Crossover}[g](\tilde{\Theta}[g]; \mi{score}_M)$}
          \WHILE{not $\textsc{StopCriterion}[g](\tilde{\Theta}[g])$}
          \STATE{$\tilde{\Theta}[g] \sim \textsc{Mutate}[g](\tilde{\Theta}[g]; \mi{score}_M)$}
          \ENDWHILE
        \ENDFOR
        \STATE{$\Theta_t \gets \tilde{\Theta}[K] \cup \tilde{\Theta}[R]$}
        \STATE{$\Gamma_t \gets \{ \underbrace{K, \ldots , K}_{ | \tilde{\Theta}[K]|~\text{copies of}~K } , \quad \underbrace{R, \ldots, R}_{ |\tilde{\Theta}[R]|~\text{copies of}~R }\}$}
        \STATE{$L_t \gets | \tilde{\Theta}[K]|+  |\tilde{\Theta}[R]| $}
        \ENDLOOP
    \end{algorithmic}
\end{algorithm}

\paragraph{\textbf{Reweight}}
This step is trivial: we simply use $\mi{score}_M$ to compute the weight of each sample.

\paragraph{\textbf{Migrate}}
This step is introduced by the dual-strategy approach.
As discussed in \cref{Se:SMCFuzzingForWCA}, we take inspiration from reproduction
strategies of organisms~\cite{kn:AH86}:
the $K$-strategy works in crowded environments and centers all the resources
on a limited number of offspring, hoping to generate high-quality offspring,
whereas the $R$-strategy works in uncrowded environments and produces a considerable
number of offspring, hoping to increase diversity.
In terms of optimization and computational resources, we realize the $K$-strategy
as \emph{exploitation}, generating a few new samples and trying to reach local optima
around the current samples; whereas we realize the $R$-strategy as \emph{exploration},
generating a lots of diverse new samples and trying to reach a high variance in terms
of their fitness scores.
In our implementation, we further split the $K$ group into an extreme-$K$ group and
a mild-$K$ group. 
The purpose of setting a-mild $K$ group is just for controlling the total number of
population;
it uses essentially the same strategy as that of the extreme-$K$ group, and is only
treated differently during migration.

\framework{} allows a random migration routine $\textsc{Migrate}$ to update group
assignments based on the current assignments and weights of the samples.
In our implementation, the migration takes place freely between the extreme-$K$ group
and the mild-$K$ group, but in a restricted manner between the extreme-$K$ group and the $R$ group.
The main idea is to build a nature-inspired routine of attracting candidates from the $R$ group to
the $K$ group, as well as eliminating those candidates with relatively small weights from the $K$ group.
Note that the motivation that drives candidates from the $R$ group to the $K$ group is greatly
influenced by the difference in average weights.
In our implementation, we use
$-\frac{\text{average weight of $K$ group} - \text{average weight of $R$ group}}{\text{largest weight}}$
to determine the migration rate from $R$ to $K$.
After we determine the migration rate, the fitness scores of the candidates are view as criteria in deciding 
their weights in a Roulette wheel selection process, which means that candidates with higher scores have higher 
chances to migrate. 
The migration rate from $K$ to $R$ is simply a fixed rate, just for discarding candidates with relatively small weights
from the $K$ group.

\paragraph{\textbf{Resample}}
This step is basically the same as the adaptive resampling step of resample-move SMC.
The difference is that \framework{} does not guarantee that the population size is a constant;
instead, the population size at the $t$th epoch is denoted as $L_t$.
This is because crossover and mutation operations can generate a different number of new samples.
Therefore, we also perform a resampling step when the current sample size is greater than
a pre-specified threshold $L_\m{max}$.

After the resampling step, our algorithm again separates the current population into two,
denoted by $\tilde{\Theta}[K]$ and $\tilde{\Theta}[R]$, for the $K$ group and the $R$ group, respectively.
For each group, the algorithm then proceeds with the rejuvenating step, which first performs
crossover-based MCMC transitions and then repeatedly performs mutation-based
MCMC transitions until some stop criterion is met.
Note that crossover, mutation, and stop criterion are all group-specific.

To recast genetic operations in our \framework{} framework, we follow the idea of
population-based MCMC~\cite{Biometrika:JSH07} and evolutionary MCMC~\cite{EA:DT03}
to integrate Metropolis-Hastings algorithms with genetic proposal distributions.
The idea has been reviewed in \cref{Se:SMCFuzzingForWCA}; thus, in the rest of this section,
we present the genetic operations used in our implementation.

\paragraph{\textbf{Crossover}}
This step is basically to carry out random uniform crossover within groups. 
Parent pairs are selected from the same group randomly, and for each pair, we sample
a $(0|1)*$ string, where the proportion of ones corresponds to crossover rate, deciding whether 
a part of the new offspring should be taken from one of the parents.

\paragraph{\textbf{Mutate}}
This step is mainly achieved utilizing bit flip, one of the standard mutation techniques. 
Mutation is performed on every candidate as a bounded exploration mechanism within a small neighborhood
of that candidate.
Every bit of the candidate is considered separately: with some probability, the bit gets flipped. 
After all bits have been decided, one mutated candidate is produced.

The setup of stopping criteria for $K$ and $R$ groups are slightly more sophisticated.
To implement the dual strategy, 
we set different stopping criteria for different groups. For the $K$ group, the computational resource is 
sufficient enough for carrying out kilos of attempts and severe competition. Therefore, the criterion is 
set to be achieving better fitness scores than surrounding candidates that are slightly different 
from the one being mutated. On the other hand, for the $R$ group, the computational resource is scarce, and only capable of 
carrying out dozens of attempts. Thus, the criterion is set to be having higher mutation potential, 
which, in our implementation, means that the standard deviation of the fitness scores of surrounding candidates
should be as high as possible. 

\paragraph{\textbf{Implementation}}
We implemented a prototype of \framework{} in Java, which consists of about 3,100 lines of code.
We have not yet implemented a instrumentation routine, so currently we assume the program-to-be-analyzed
is implemented as a Java class with an entry method and the program includes explicit tick statements to
indicate resource usages (like the example in \cref{Fi:RunningExample}(a)).
In some of the evaluation subjects, we actually utilized AFL~\cite{misc:AFL} to carry out instrumentation.
The source code of our prototype and the programs used in our evaluation
are included in the submitted replication package.

\section{Experimental Evaluation}
\label{Se:Evaluation}

In this section, we present an experimental evaluation of our implementation of \framework{}, in the hope of answering the following research questions. 
\begin{itemize}
    \item \textbf{RQ1:} How well does \framework{} perform in comparison with existing black-box worst-case-analysis tools?
    \item \textbf{RQ2:} Does the \framework{}'s MCMC-based rejuvenation outperform a locally-optimal variant of \framework{}?
    % \item $\mathbf{RQ3:}$ In what degree may we rely on the model to generate the worst-case input for further analysis?
\end{itemize}

% In order to make our contribution available for public use, we have released the full version of our presented model and evaluation artifacts on Github:
% https://github.com/Aaronlosl/GenericUrbanization.

\subsection{Experimental Setup}
\label{Se:Setup}

To present a versatile evaluation of how \framework{} may have an impact on both the basic algorithms and complex industrial applications, 
we select eight evaluation subjects, summarized in \cref{Ta:Evaluation}.
Some of the subjects are adapted from recent work on WCA, e.g., \slowfuzz{}~\cite{CCS:PZK17} and \mayhem{}~\cite{SP:CAR12}.
The subjects on the left column of \cref{Ta:Evaluation} are basic algorithms that may serve to create a general sense of how \framework{} works in terms of input mutation and worst-case navigating. 
On the other hand, the subjects on the right column of \cref{Ta:Evaluation} are believed to be of great importance to the software industry. 
Note that we use different resource metrics for different subjects. 
The subjects on the left column are to be evaluated by observing the steps that the algorithm takes to halt, whereas 
those on the right column via counting the jumps that the application takes to complete the task.

\begin{table}
\centering
    \caption{Overview of the evaluation subjects}
        \begin{tabular}{||c|c||c|c||}
            \hline
            \textbf{ID} & \textbf{\textit{Subject}}& \textbf{ID}& \textbf{\textit{Subject}} \\
            \hline
            1 & Generate ordered pairs & 5 & RegEx$^{\mathrm{a}}$ \\
            2 & Insertion sort & 6 & Hash table \\
            3 & Quicksort & 7 & Compression \\
            4 & Tree sort & 8 & Smart contract \\
            \hline
            \multicolumn{4}{l}{$^{\mathrm{a}}$with fixed regex expression.}
        \end{tabular}
        \label{Ta:Evaluation}
\end{table}

For all evaluation subjects, we ran four WCA tools:
(1) \kelinci{}, (2) \kelinciWCA{}, (3) Locally-optimal \framework{}, and (4) \framework{}.
\kelinci{} provides an interface for running \afl{}~\cite{misc:AFL} on Java programs~\cite{CCS:KLP17}
and \kelinciWCA{} extends \kelinci{} to prioritize execution paths with high resource consumption~\cite{ISSTA:NKP18}.
We include in the comparison a \emph{locally-optimal} variant of \framework{},
which replaces the MCMC-based rejuvenation step with a local search that only keeps a locally optimal new candidate.
It is of vital importance to control the basic setup and the initial value as they can greatly affect the performance of each tool. 
Therefore, we use the same meaningless inputs and default parameters to setup our evaluation.
We ran each tool on each subject for at most 100 epochs or 100 minutes,
which we will elaborate in the setup of each evaluation subject.

Note that there are also recently proposed WCA tools, such as \slowfuzz{}~\cite{CCS:PZK17}, that execute evolutionary WCA on binary codes.
Because we setup our experiments on Java programs, we cannot compare \framework{} with them directly.
Nevertheless, we find that \badger{}~\cite{ISSTA:NKP18}, a WCA tool that integrates fuzzing and symbolic execution,
can serve as an indirect indicator for comparison. 
The authors of \badger{} claim that black-box fuzzing-based WCA tools like \slowfuzz{} are in spirit similar
to \kelinciWCA{}.
Thus, our comparison with \kelinciWCA{} should reflect how our \framework{} framework performs against \slowfuzz{}, etc.
A more systematic comparison (e.g., reimplement \slowfuzz{} and \mayhem{}'s algorithms for analyzing Java programs)
is left for future work.

All of our experiments were conducted on an x86-64 architecture, Intel Core i7-1365UE 4.9GHz Linux machine with 32 GB of memory. 
We used \jdk{} 1.9.0\_132 and configured the Java VM to use at most 10 GB of memory.

\subsection{Ordering}
We embark on the very basic algorithm in computer science: comparing and ordering. 
The first experiment we conducted is on pairing procedures to collect pairs that satisfy specific conditions, or in the case of our setup, ordered adjacent pairs. 
Because the structure of the experiment is relatively less sophisticated, the result shows that our approach behaves significantly better than \kelinci{} and \kelinciWCA{}.

\begin{figure}[h]
    \centering
%	\begin{minipage}[t]{1.0\linewidth}  \label{Fig.3} 
%		\hspace{24mm}
		\includegraphics[width=8.76cm,height=5.16cm]{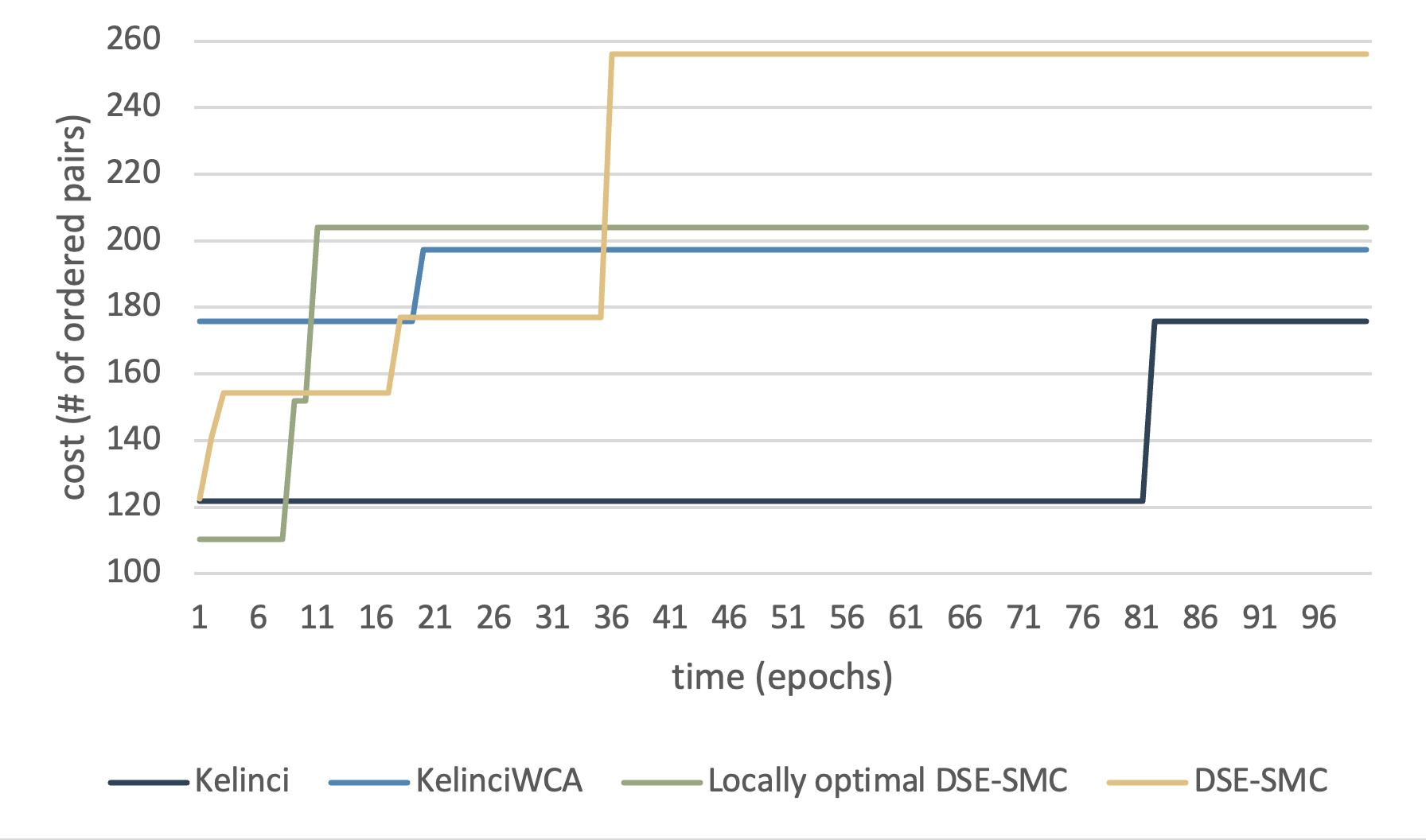}
		\caption{Result for generating ordered pairs ($N=20$)}
		\label{Fi:Ordering}
%	\end{minipage}
\end{figure}

\cref{Fi:Ordering} displays the result of generating ordered pairs on a given array, where $N$
is the length of the array.
The exact maximum score in this task is $269$, and \framework{} reaches $95\%$ worst-case performance in less than $40$ minutes. 
All of the evaluated tools reach scores no less than $65\%$ of the maximum score in $100$ epochs, and converge no greater than $80$ epochs. 

\subsection{Sorting}
\begin{figure}[h]
    \centering
%	\begin{minipage}[t]{1.0\linewidth}  \label{Fig.4} 
%		\hspace{24mm}
		\includegraphics[width=8.76cm,height=5.16cm]{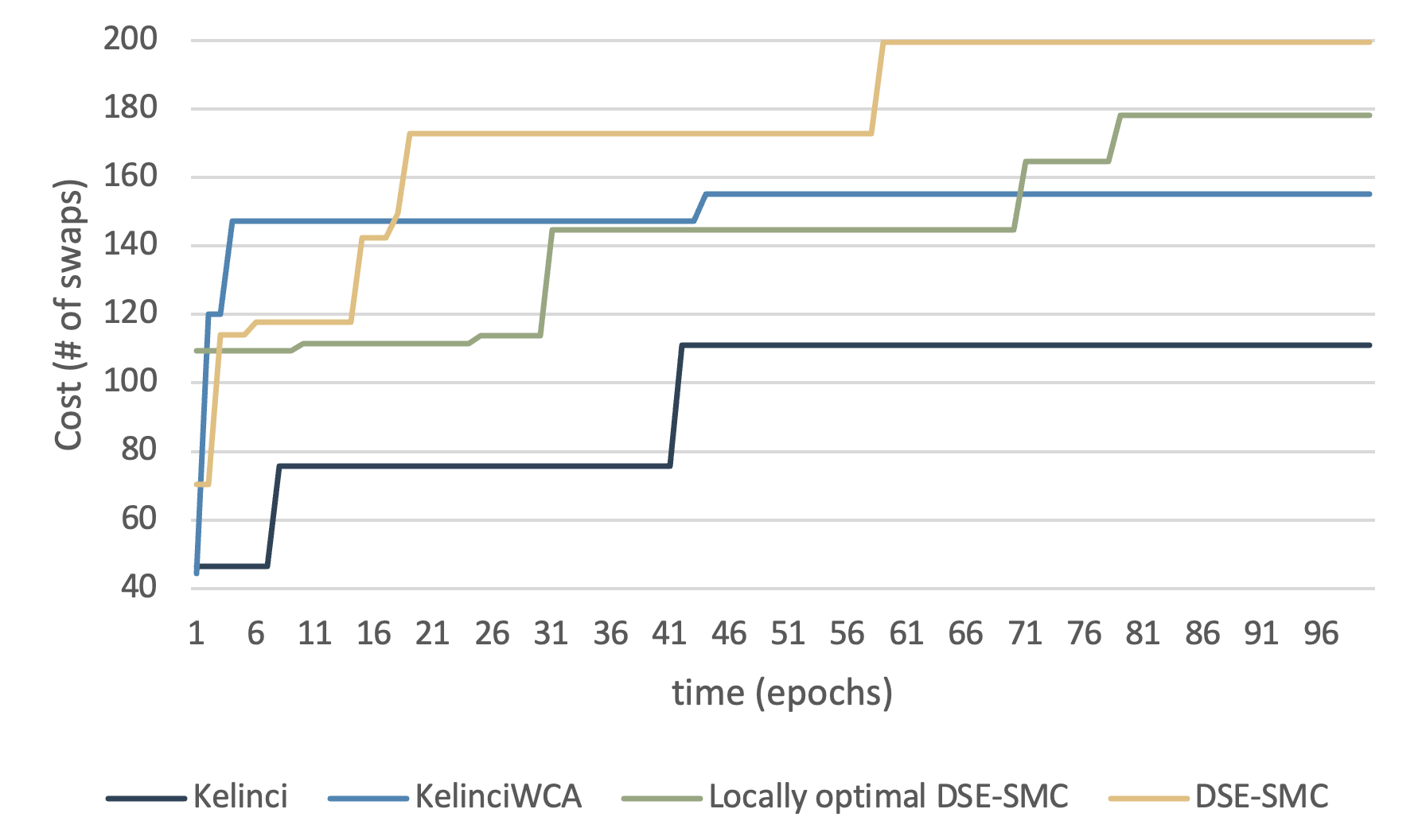}
		\caption{Result for insertion sort ($N=20$)}
		\label{Fi:InsertionSort}
%	\end{minipage}
\end{figure}

We then evaluate how our framework reacts to textbook algorithms, which are also evaluated in other WCA tools. 
In this section, we consider three sorting algorithms.
Insertion sort and quicksort vary in terms of time complexity and the number of jumps. 
Both of them are easy to implement and execute, yet the worst-case behavior may be significantly influenced by the size of the input. 
In line with the execution time limit, we conducted our experiment on an implementation of insertion sort under input size of $20$ and present our results in \cref{Fi:InsertionSort}, where $N$ is the input size. 

The exact maximum score (i.e., the number of swaps) possible for insertion sort is $289$.
From the result shown in \cref{Fi:InsertionSort}, we observe that the score increases continuously when running \framework{}. 
The use of genetic mutation and crossover generally reduce the running time consumed to converge by $60$ epochs. 
Last but not least, \kelinciWCA{} utilizes the customized cost function and indeed performed better at first $30$ epochs, yet failed to mutate towards larger scores in a restricted time frame.

\begin{figure}[h]
    \centering
%	\begin{minipage}[t]{1.0\linewidth}  \label{Fig.5} 
%		\hspace{24mm}
		\includegraphics[width=8.76cm,height=5.16cm]{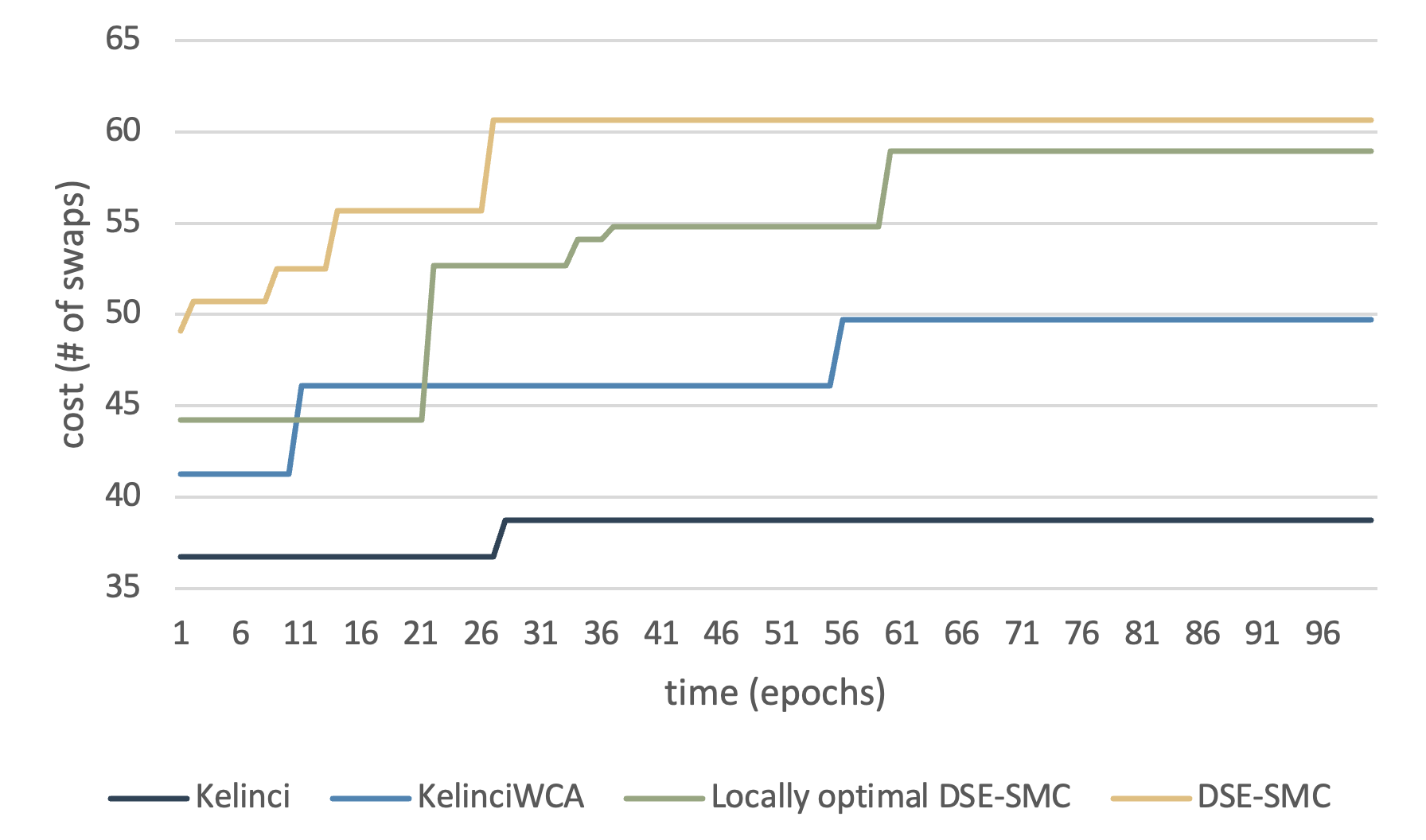}
		\caption{Result for quicksort ($N=20$)}
		\label{Fi:Quicksort}
%	\end{minipage}
\end{figure}

The results for quicksort are shown in \cref{Fi:Quicksort}.
After executing for 100 epochs, \framework{} behaves better than \kelinci{} by $20\%$. 
On the other hand, we also notice that \framework{} behaves quite unsteady on the quicksort task. 
The minor drawback may be explained by the probabilistic mutation process and its inherited randomness. 
Overall, or in a slightly longer period, \framework{} converges 8 times quicker than \kelinciWCA{}. 

\begin{figure}[h]
    \centering
%	\begin{minipage}[t]{1.0\linewidth}  \label{Fig.6} 
%		\hspace{24mm}
		\includegraphics[width=8.76cm,height=5.16cm]{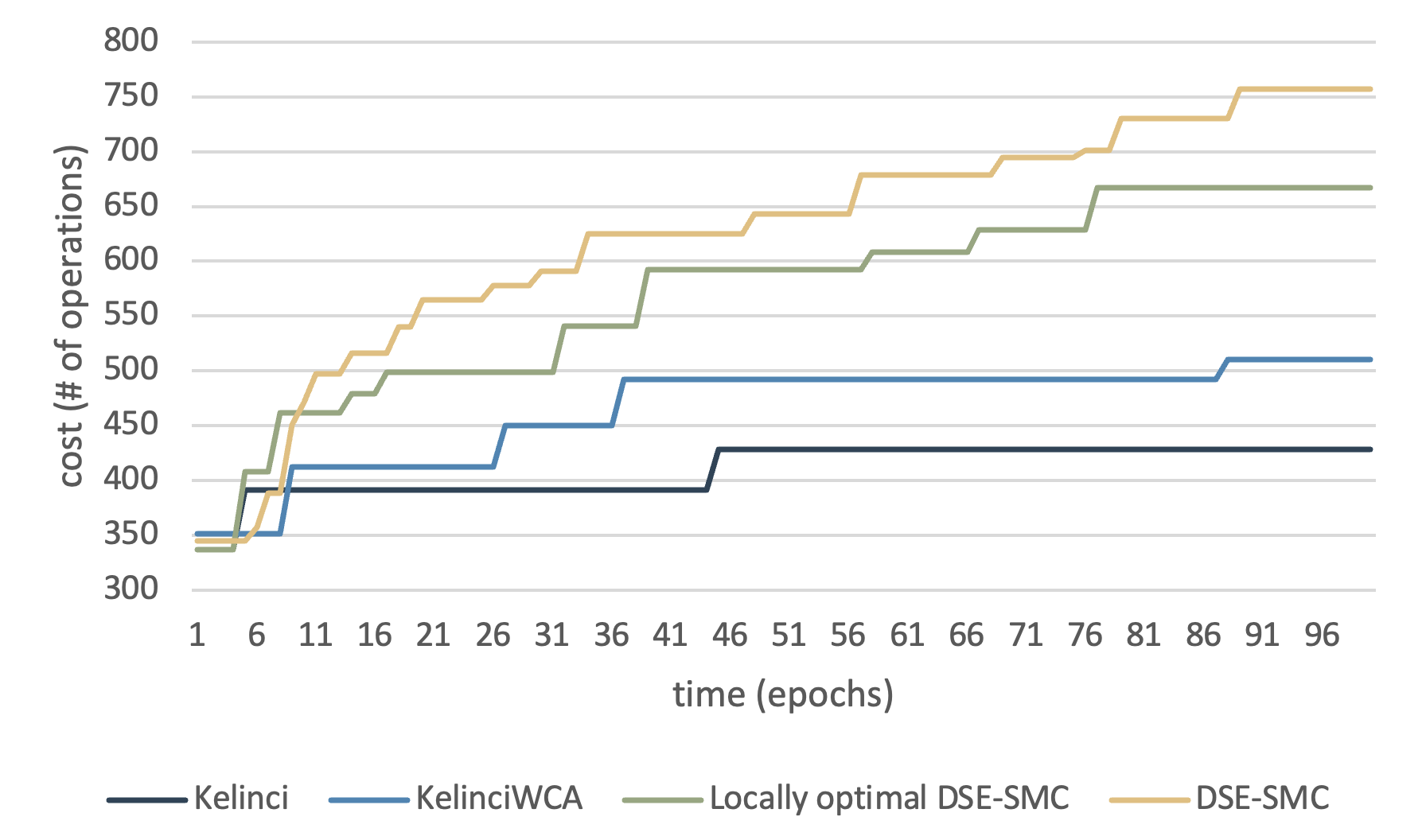}
		\caption{Result for tree sort ($N=20$)}
		\label{Fi:DFS}
%	\end{minipage}
\end{figure}

The last sorting task we investigated is the process we called tree sort.
It is used to demonstrate the performance of \framework{} on subjects that use complex data structures.
In this subject, we implemented a red-black tree for inserting integers and set the resource metric to be
the number of red-black tree operations. 
The tendency and convergence rate of \framework{} in comparison with other tools are obvious, as shown in \cref{Fi:DFS}:
our tool achieved a significantly faster speed to discover worst-case inputs.

\subsection{RegEx}
To secure an idea of what our approach behaves in extreme cases which are generally considered important in detecting vulnerabilities, we conduct the experiment on regular expression Denial-of-Service (ReDos) vulnerabilities. 
In accordance with prior work (e.g.,~\cite{ISSTA:NKP18}), we used the \verb|java.util.regex| JDK package. 
It is generally used for vulnerability revelation and not exactly suitable to test worst-case generating. 
Yet the results may be of help to understand the cons of applying black-box techniques to those type of problems. 

In this experiment, we fixed the regular expression and attempted to mutate the text for worst-case inputs. 
The ultimate goal for this example is that the WCA tools generate a password that matches the regular expression \verb|((?=.*\d)(?=.*[a - z])(?=.*[A - Z])(?=.*[@#$%]).{6, 20})|. 

\begin{figure}[h]
    \centering
%	\begin{minipage}[t]{1.0\linewidth}  \label{Fig.7} 
%		\hspace{24mm}
		\includegraphics[width=8.76cm,height=5.16cm]{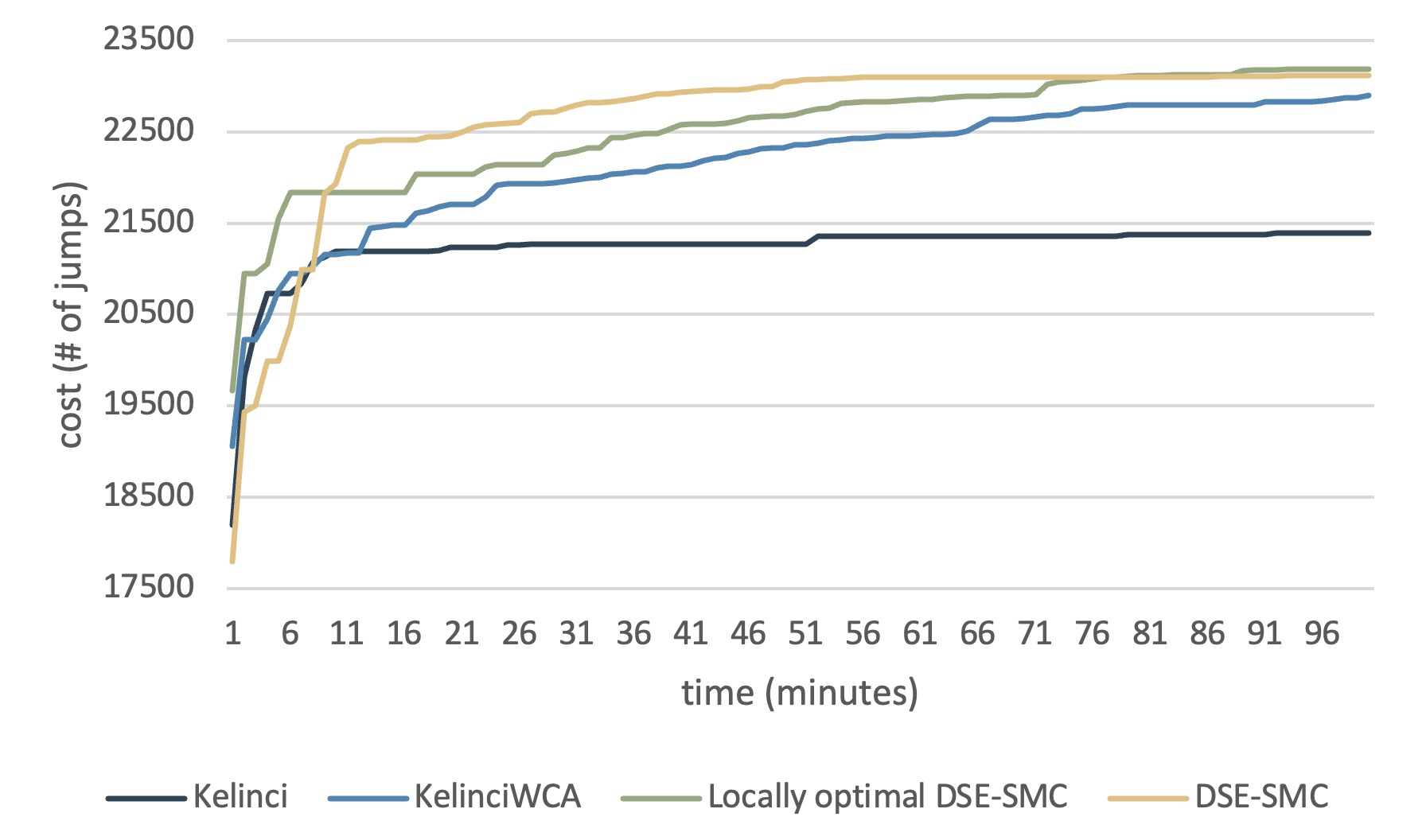}
		\caption{Result for regex}
		\label{Fi:RegEx}
%	\end{minipage}
\end{figure}

\cref{Fi:RegEx} shows the experimental result. \framework{} behaved slightly worse than the locally-optimal variant of \framework{}, and got caught up by \kelinci{} at approximately $110$ minutes. 
The results can be interpreted as lack of detailed domain knowledge with regard to the program structure. 
White- or grey-box tools in general have components for condition analysis and dynamic instrumentation. 
While black-box tools (like ours) execute faster in a single epoch, there is a choice between sacrificing computational speed and extracted information in one epoch when it comes to structural information within the program. 
Interestingly, the performance of \framework{} varies greatly across different runs.
Inherited from probabilistic computation, \framework{} displays specific characters of randomness. 

\subsection{Hash Table}
\begin{figure}[h]
    \centering
%	\begin{minipage}[t]{1.0\linewidth}  \label{Fig.8} 
%		\hspace{24mm}
		\includegraphics[width=8.76cm,height=5.16cm]{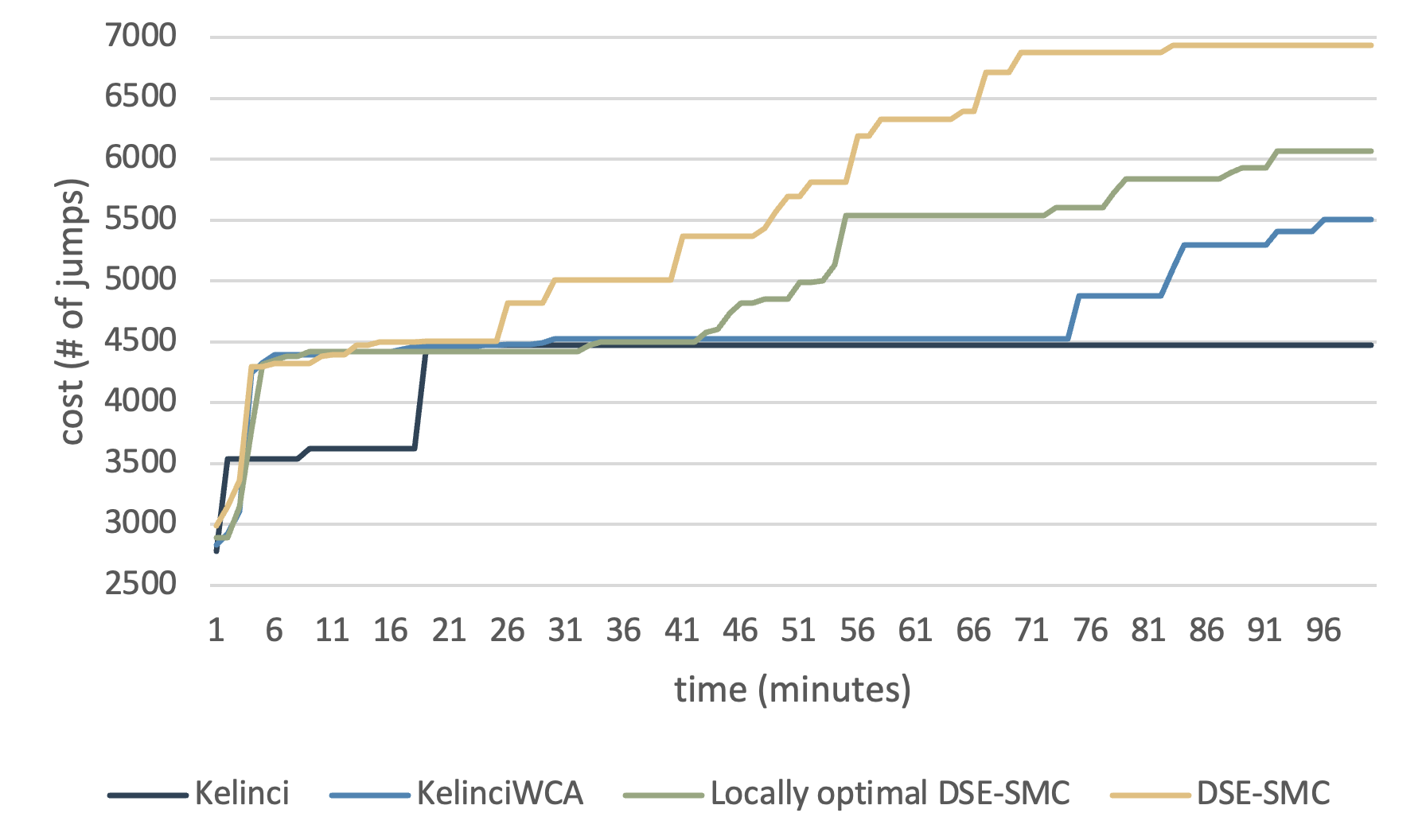}
		\caption{Result for hash table}
		\label{Fi:HashTable}
%	\end{minipage}
\end{figure}

The hash-table subject is also implemented with considerations similar to \emph{RegEx}. 
We modified the input part of the hash-table algorithm in accordance with prior work. 
Based on the outcome, we observe that there are several noises that seem to show that our tool generates worst-case inputs significantly better.  
By the one-hour time limit, our full version of \framework{} has already generated input cases that can trigger more than $6,300$ jumps, while \kelinciWCA{} and locally-optimal \framework{} got stuck at near $5,500$ jumps.
That displays a significant advancement, as shown in \cref{Fi:HashTable}.

Another observation is that our tool continues to make progress at the end of the evaluation process.
This indicates that probabilistic implementation of the genetic operations possess higher potential of exploration and the dual-strategy approach does make sense with regard to biological investigation. 
In addition, unlike the previous subjects of experiments, the performance of our tool on this subject is relatively stable in comparison, indicating the particular posterior distribution is a relatively smooth territory.

\subsection{Compression}
A normal application of WCA is to analyze the performance of compression algorithms. 
Compression algorithms are a hot topic in both the industry and the academia.
We used the \verb|org.apache.commons.compress| JDK package to conduct our experiment. 
The subject to be tested is the performance of compression algorithms on $100$ bytes of input data.

\begin{figure}[h]
    \centering
%	\begin{minipage}[t]{1.0\linewidth}  \label{Fig.9} 
%		\hspace{24mm}
		\includegraphics[width=8.76cm,height=5.16cm]{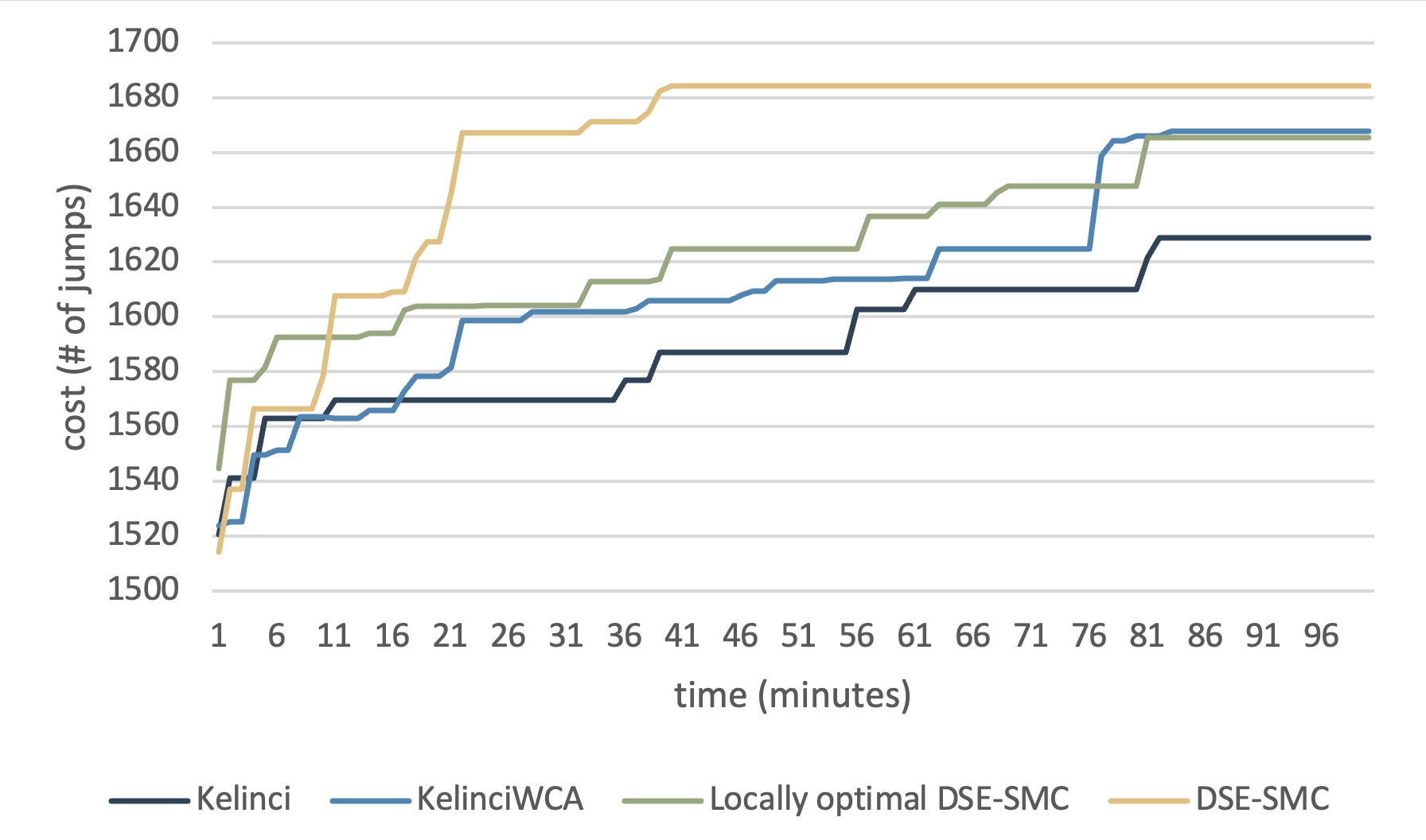}
		\caption{Result for compression}
		\label{Fi:Compression}
%	\end{minipage}
\end{figure}

\cref{Fi:Compression} shows the experimental result.
Despite the difference in initial performances, \framework{} reaches 1680 jumps at least 3 times faster than \kelinciWCA{}, and the locally optimal \framework{} makes steady progress and performs slightly better than \kelinciWCA{}. 
The locally-optimal variant of \framework{} generated initial inputs that already lead to relatively high resource costs, yet slowed down quickly and got caught up by \kelinciWCA{} at 23 minutes. 
\framework{}, on the other hand, reveals great potential in \emph{exploration} that it boosts quickly, for example, in the $11$th and $21$th minutes. 
This can be explained by the fact that we use genetic mutation loops, which adds more steps relevant to exploration. 
It is also worth mentioning that we observe a slowdown after the boosts, indicating that our framework probably reaches a local optimum and bit-level mutation does no longer lead to better exploration. 
Although \framework{} continues to make progress after 100 minutes, at that point, one should decrease the rate of annealing the population and the migration rate of candidates applying the $R$-strategy to address the issue and look for stabler balance between exploration and exploitation. 

\subsection{Smart Contract}

One of the most interesting potentials for worst-case estimation is the potential of gas fuzzing in modern blockchains. 
Generally, the gas is a deterministic indicator that determines the cost of a smart contract and the resource consumption in executing it. 
We setup the environment in a simple smart contract in a Java-based form. Details can be found in the replication package.

\begin{figure}[h]
    \centering
%	\begin{minipage}[t]{1.0\linewidth}  \label{Fig.10} 
%		\hspace{24mm}
		\includegraphics[width=8.76cm,height=5.16cm]{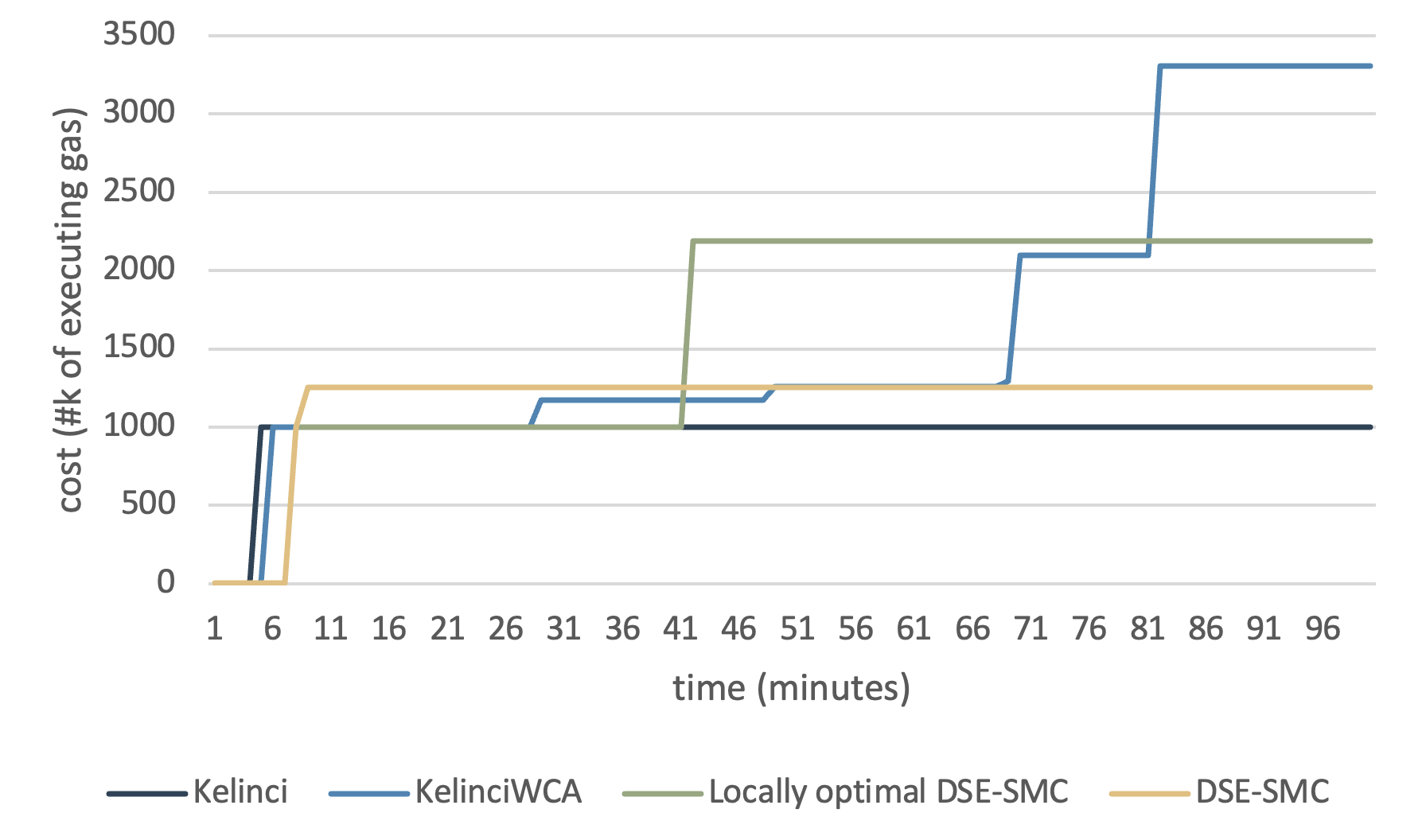}
		\caption{Result for smart contract}
        \label{Fi:SmartContract}		
%	\end{minipage}
\end{figure}

Results for this subject are displayed in \cref{Fi:SmartContract}.
As the cost is user-defined, we have to manually set the cost in order to feed \kelinci{} and \kelinciWCA{}.
We conducted the experiment for multiple times and got results with huge variation. 
Our \framework{} tool does not perform as good as one would expect. 
An observation from the smart-contract experiment is that \framework{} generally got stuck at 10 minutes and behaved worse than \kelinciWCA{}. 
In addition, time consumed in one epoch is relatively greater than other tools due to the implementation of different strategies and relatively complex rejuvenation routines. 
Note that the smart-contract experiment we conduct is based on its gas consumption, whose calculation mainly replies on memory allocation and pointers' inner relationships. 
Without inner relationship or fine-grained program analysis, black-box based tools are easy to miss target branches which consume a lot of gas, and therefore, fail to observe overall worse-case gas-usage behavior. 
It would be interesting future research to integrate our black-box technique with white- or grey-box approaches that can
provide more application-specific information.

\subsection{Threats to Validity}

\paragraph{Internal Validity}
The main threat to internal validity is the theoretical structure of \framework{}. 
We have cross validate the revise the theoretical process to ensure the solidity as best as we can. 
Another threat worth mentioning is the choice of parameters that are integrated in our framework. 
As it is computational costly, and therefore, impractical for testers to manually set those parameters to suit each subjects, we did pre-experiments beforehand on subject 2 to mitigate the effects. 
Yet the choice can still be customized, and consequently display results significantly different from each other to concernable extent. 
Also, experiments are conducted on server that are not often reliable in terms of execution speed and resource consumption. 
In future practice, we will look into methods of parallelizing the dual strategy implementation. 
Theoretical speaking, we could expect a speedup of 1.5x, which is an outstanding improvement in efficiency. 

\paragraph{External Validity}
The main threat to external validity is the direction in which we pick the subjects and conduct the experiments. 
These subjects are not generalized or regulated, and therefore may not be suitable for tool comparing to some related work. 
To mitigate this concern, we select many of our benchmarks that are in line with the existing work similar to ours. 
In addition, the pile insertion part for basic performance monitoring of our framework is implemented using external tools, which can potentially be incorrect and threaten our validity. 

\section{Related Work}
\label{Se:Related}

\paragraph{Worst-Case Analysis}
Most related to our work are techniques for conducting worst-case analysis by generating
worst-case inputs.
In this paper, we focus on black-box approaches, which are agnostic of application-specific
or implementation-specific domain knowledge.
Most recent work along this line includes \kelinciWCA{}~\cite{ISSTA:NKP18} and \slowfuzz{}~\cite{CCS:PZK17}.
We have presented an empirical comparison of our \framework{} framework to those techniques in \cref{Se:Evaluation}.
The major difference is that \framework{} is based on an WCA-is-MAP observation and a novel dual-strategy
evolutionary sequential-Monte-Carlo algorithm.

Another popular line of work for worst-case analysis is based on symbolic execution.
Those techniques need to look into the concrete structures of analyzed programs, so they
are fundamentally different from our focus in this paper.
WISE~\cite{ICSE:BJS09} explores all program paths to find worst-case ones on small inputs,
and then uses those paths to derive a heuristic to generate larger worst-case inputs.
SPF-WCA~\cite{ICST:LKP17} introduces path policies to guide the search of symbolic execution.
\citet{POPL:WH19} proposes a type-guided worst-case input generation that utilizes
resource-aware type derivation~\cite{POPL:HDW17} to prune the search space of symbolic execution.
\badger{}~\cite{ISSTA:NKP18} combines symbolic execution with fuzzing techniques
to generate high-resource-usage inputs to avoid exhaustive exploration of program paths.
Our future research may investigate how to incorporate symbolic execution in \framework{}.

\paragraph{Fuzzing}
Fuzzing receives more and more attraction now that the scale of codes increases significantly~\cite{misc:MYG12}.
It has the following advantages over other testing techniques: (1) no requirements for source code of the analyzed program, (2) more faster in comparison to human testers, (3) relatively less expensive, and (4) more generic and portable.
In our work, \framework{} can be seen as a fuzzer that targets the generation of worst-case inputs for Java programs.
The tool is conceptually similar to general black-box fuzzers including \afl{}~\cite{misc:AFL} and \libfuzzer{}~\cite{misc:LibFuzzer}.
Our innovation is the integration of evolutionary fuzzing into the sequential-Monte-Carlo framework and the development
of the dual-strategy approach.

Similar to WCA, recent fuzzers also possesses great potential especially with the help of symbolic-execution engines. 
To name a few,
\mayhem{}~\cite{SP:CAR12}, a symbolic execution engine, combined with the \murphy{} fuzzer, won the 2016 DARPA Cyber Grand Challenge. 
\evosuite{}~\cite{ISSRE:GFA13} is a test-case generation tool with dynamic symbolic execution for Java. 

\paragraph{Sequential Monte Carlo}
The key sequencial structure of \framework{} is based on sequential Monte Carlo (SMC). 
There are many developed algorithms that can be integrated in \framework{} as a sequence building techniques, including SMC samplers~\cite{JRSS:DDJ06}, bootstrap SMC filters~\cite{SPM:Candy07}, resample-move SMC~\cite{JRSS:GB01,Biometrika:Chopin02}, etc.
\citet{book:CSI2000} summarizes some commonly used cases of SMC.
Despite some concerns on degeneracy~\cite{SPM:Candy07} and introducing non-independent data, bridge sampling~\cite{JCP:Bennett76} alleviates the concerns of cases with high-dimensional posterior (e.g.~\cite{TEJ:FS04}). 
Generally, SMC serves as a crucial part in dynamic bayesian models, for example, MCMC rejuvenation~\cite{JRSS:GB01}.
SMCP${}^3$~\cite{AISTATS:LMZ23} has broaden the scope of SMC to incorporate probabilistic auxiliary variables during inference. 
Besides, if the prior and posterior distributions have a similar shape or strong overlap, a naive Monte Carlo estimator could also be a satisfactory choice.
Starting from this work, we hope to systematically investigate SMC-based fuzzing for software testing.

%\paragraph{Probabilistic Programming}
%\framework{} supports probabilistic algorithms as \emph{mutation} and \emph{testing}, therefore, yield greater potential in automation, computational consumption as well as model performance.
%Since the proposal of probabilistic development kits, including Pyro (TODO), probabilistic programming language (PPL) has been greatly explored. 
%SMC and MCMC are supported by various models (TODO). 
%In addition, many modern applications in artificial intelligence take advantage of PPL for faster speed (TODO). 

\paragraph{Nature-Inspired Algorithms}
The population-based meta-heuristics approach we adapt in this paper is to address the concern that single-solution-based meta-heuristics may get stuck in local optima~\cite{AES:KK17}.  
The process connecting the sequential structure within \framework{} is the originated from genetic algorithms~\cite{book:Goldberg89}, and inspired by swarm intelligence algorithms like particle swarm optimization (PSO)~\cite{ICNN:KE95} ant colony optimization (ACO)~\cite{ANTS08}.
%Optimizers apply genetic algorithms consequently and reveals great potential, including ... (TODO). 
%Not only in the field of software testing, but a great variety of computation-based optimization tasks devices genetic algorithms for sequence sampling (TODO). 
Intuitively and naturally, nature-inspired algorithms have been put forward to imitate, simulate and better approximate the natural process of evolution in complicated environment than classic algorithms.
It would be interesting future research to incorporate more nature-inspired algorithms into sequential Monte Carlo.
\section{Conclusion}
\label{Se:Conclusion}

We have presented \underline{D}ual-\underline{S}trategy \underline{E}volutionary Sequential Monte-Carlo (\framework{}) framework for black-box worst-case-analysis fuzzing.
The framework is built upon a key observation discussed in this paper: worst-case analysis (WCA) is isomorphic to maximum-a-posteriori (MAP) estimation in the context of Bayesian inference.
\framework{} incorporates resample-move SMC, evolutionary MCMC, nature-inspired $K$- and $R$-strategies, and fuzz testing to provide a generic, adaptive, and sound framework for worst-case analysis.
We implemented a prototype of \framework{} and evaluated its effectiveness on several subject programs.

In the future, we plan to dig further into the correspondence between WCA and MAP.
So far we have not used the \emph{prior} distributions that are common in probabilistic models.
In the context of Bayesian inference, prior distributions are usually used to encode domain
information, and thus to guide inference algorithms.
Is it possible to cast white-box analysis results (e.g., program analysis, type derivation, and symbolic execution)
as prior distributions to provide insights for WCA?
Another research direction is to transfer other Bayesian-inference algorithms to carry out WCA.
For example, recent advances in variational inference render it as a promising technique for approximating
posterior distributions.
It would be interesting to see if it is applicable to WCA.
%\input{dataavailability}

%\begin{acks}
%To Robert, for the bagels and explaining CMYK and color spaces.
%\end{acks}

\bibliographystyle{ACM-Reference-Format}
\bibliography{lit,db}


\begin{thebibliography}{49}

%%% the Web address.

\ifx \showCODEN    \undefined \def \showCODEN     #1{\unskip}     \fi
\ifx \showDOI      \undefined \def \showDOI       #1{#1}\fi
\ifx \showISBNx    \undefined \def \showISBNx     #1{\unskip}     \fi
\ifx \showISBNxiii \undefined \def \showISBNxiii  #1{\unskip}     \fi
\ifx \showISSN     \undefined \def \showISSN      #1{\unskip}     \fi
\ifx \showLCCN     \undefined \def \showLCCN      #1{\unskip}     \fi
\ifx \shownote     \undefined \def \shownote      #1{#1}          \fi
\ifx \showarticletitle \undefined \def \showarticletitle #1{#1}   \fi
\ifx \showURL      \undefined \def \showURL       {\relax}        \fi
% invisible to TeX
\providecommand\bibfield[2]{#2}
\providecommand\bibinfo[2]{#2}
\providecommand\natexlab[1]{#1}
\providecommand\showeprint[2][]{arXiv:#2}

\bibitem[Andrews and Harris(1986)]%
        {kn:AH86}
\bibfield{author}{\bibinfo{person}{John~H. Andrews} {and} \bibinfo{person}{Robin~F. Harris}.} \bibinfo{year}{1986}\natexlab{}.
\newblock \showarticletitle{{$r$- and $K$-Selection and Microbial Ecology}}.
\newblock In \bibinfo{booktitle}{\emph{Advances in Microbial Ecology}}. \bibinfo{publisher}{Springer, Boston, MA}.
\newblock
\urldef\tempurl%
\url{https://doi.org/10.1007/978-1-4757-0611-6_3}
\showDOI{\tempurl}


\bibitem[Barthe et~al\mbox{.}(2020)]%
        {book:FoundProbProg20}
\bibfield{editor}{\bibinfo{person}{Gilles Barthe}, \bibinfo{person}{Joost-Pieter Katoen}, {and} \bibinfo{person}{Alexandra Silva}} (Eds.). \bibinfo{year}{2020}\natexlab{}.
\newblock \bibinfo{booktitle}{\emph{{Foundations of Probabilistic Programming}}}.
\newblock \bibinfo{publisher}{Cambridge University Press}.
\newblock
\urldef\tempurl%
\url{https://doi.org/10.1017/9781108770750}
\showDOI{\tempurl}


\bibitem[Bennett(1976)]%
        {JCP:Bennett76}
\bibfield{author}{\bibinfo{person}{Charles~H Bennett}.} \bibinfo{year}{1976}\natexlab{}.
\newblock \showarticletitle{{Efficient estimation of free energy differences from Monte Carlo data}}.
\newblock \bibinfo{journal}{\emph{J. Comput. Phys.}}  \bibinfo{volume}{22} (\bibinfo{date}{October} \bibinfo{year}{1976}), \bibinfo{pages}{245--268}.
\newblock
Issue 2.
\urldef\tempurl%
\url{https://doi.org/10.1016/0021-9991(76)90078-4}
\showDOI{\tempurl}


\bibitem[Bingham et~al\mbox{.}(2018)]%
        {JMLR:BCJ18}
\bibfield{author}{\bibinfo{person}{Eli Bingham}, \bibinfo{person}{Jonathan~P. Chen}, \bibinfo{person}{Martin Jankowiak}, \bibinfo{person}{Fritz Obermeyer}, \bibinfo{person}{Neeraj Pradhan}, \bibinfo{person}{Theofanis Karaletsos}, \bibinfo{person}{Rishabh Singh}, \bibinfo{person}{Paul Szerlip}, \bibinfo{person}{Paul Horsfall}, {and} \bibinfo{person}{Noah~D. Goodman}.} \bibinfo{year}{2018}\natexlab{}.
\newblock \showarticletitle{{Pyro: Deep Universal Probabilistic Programming}}.
\newblock \bibinfo{journal}{\emph{J.\ Machine Learning Research}}  \bibinfo{volume}{20} (\bibinfo{date}{January} \bibinfo{year}{2018}).
\newblock
Issue 1.
\urldef\tempurl%
\url{https://dl.acm.org/doi/10.5555/3322706.3322734}
\showURL{%
\tempurl}


\bibitem[Borgstr{\"o}m et~al\mbox{.}(2016)]%
        {ICFP:BLG16}
\bibfield{author}{\bibinfo{person}{Johannes Borgstr{\"o}m}, \bibinfo{person}{Ugo~Dal Lago}, \bibinfo{person}{Andrew~D. Gordon}, {and} \bibinfo{person}{Marcin Szymczak}.} \bibinfo{year}{2016}\natexlab{}.
\newblock \showarticletitle{{A Lambda-Calculus Foundation for Universal Probabilistic Programming}}. In \bibinfo{booktitle}{\emph{Int.\ Conf.\ on Functional Programming}} \emph{(\bibinfo{series}{ICFP'16})}.
\newblock
\urldef\tempurl%
\url{https://doi.org/10.1145/2951913.2951942}
\showDOI{\tempurl}


\bibitem[Burnim et~al\mbox{.}(2009)]%
        {ICSE:BJS09}
\bibfield{author}{\bibinfo{person}{Jacob Burnim}, \bibinfo{person}{Sudeep Juvekar}, {and} \bibinfo{person}{Koushik Sen}.} \bibinfo{year}{2009}\natexlab{}.
\newblock \showarticletitle{{WISE: Automated Test Generation for Worst-Case Complexity}}. In \bibinfo{booktitle}{\emph{Int.\ Conf.\ on Softw.\ Eng.}} \emph{(\bibinfo{series}{ICSE'09})}. \bibinfo{pages}{463--473}.
\newblock
\urldef\tempurl%
\url{https://doi.org/10.1109/ICSE.2009.5070545}
\showDOI{\tempurl}


\bibitem[Candy(2007)]%
        {SPM:Candy07}
\bibfield{author}{\bibinfo{person}{James~V. Candy}.} \bibinfo{year}{2007}\natexlab{}.
\newblock \showarticletitle{{Bootstrap Particle Filtering}}.
\newblock \bibinfo{journal}{\emph{Signal Processing Magazine}}  \bibinfo{volume}{24} (\bibinfo{date}{July} \bibinfo{year}{2007}), \bibinfo{pages}{73--85}.
\newblock
Issue 4.
\urldef\tempurl%
\url{https://doi.org/10.1109/MSP.2007.4286566}
\showDOI{\tempurl}


\bibitem[Carpenter et~al\mbox{.}(2017)]%
        {JSS:CGH17}
\bibfield{author}{\bibinfo{person}{Bob Carpenter}, \bibinfo{person}{Andrew Gelman}, \bibinfo{person}{Matthew~D. Hoffman}, \bibinfo{person}{Daniel Lee}, \bibinfo{person}{Ben Goodrich}, \bibinfo{person}{Michael Betancourt}, \bibinfo{person}{Marcus Brubaker}, \bibinfo{person}{Jiqiang Guo}, \bibinfo{person}{Peter Li}, {and} \bibinfo{person}{Allen Riddell}.} \bibinfo{year}{2017}\natexlab{}.
\newblock \showarticletitle{{Stan: A Probabilistic Programming Language}}.
\newblock \bibinfo{journal}{\emph{J.\ Statistical Softw.}}  \bibinfo{volume}{76} (\bibinfo{date}{January} \bibinfo{year}{2017}).
\newblock
Issue 1.
\urldef\tempurl%
\url{https://doi.org/10.18637/jss.v076.i01}
\showDOI{\tempurl}


\bibitem[Cha et~al\mbox{.}(2012)]%
        {SP:CAR12}
\bibfield{author}{\bibinfo{person}{Sang~Kil Cha}, \bibinfo{person}{Thanassis Avgerinos}, \bibinfo{person}{Alexandre Rebert}, {and} \bibinfo{person}{David Brumley}.} \bibinfo{year}{2012}\natexlab{}.
\newblock \showarticletitle{{Unleashing Mayhem on Binary Code}}. In \bibinfo{booktitle}{\emph{Symposium on Security and Privacy}} \emph{(\bibinfo{series}{SP'12})}. \bibinfo{pages}{380--394}.
\newblock
\urldef\tempurl%
\url{https://doi.org/10.1109/SP.2012.31}
\showDOI{\tempurl}


\bibitem[Chen et~al\mbox{.}(2000)]%
        {book:CSI2000}
\bibfield{author}{\bibinfo{person}{Ming-Hui Chen}, \bibinfo{person}{Qi-Man Shao}, {and} \bibinfo{person}{Joseph~G. Ibrahim}.} \bibinfo{year}{2000}\natexlab{}.
\newblock \bibinfo{booktitle}{\emph{{Monte Carlo Methods in Bayesian Computation}}}.
\newblock \bibinfo{publisher}{Springer New York, NY}.
\newblock
\urldef\tempurl%
\url{https://doi.org/10.1007/978-1-4612-1276-8}
\showDOI{\tempurl}


\bibitem[Chopin(2002)]%
        {Biometrika:Chopin02}
\bibfield{author}{\bibinfo{person}{Nicolas Chopin}.} \bibinfo{year}{2002}\natexlab{}.
\newblock \showarticletitle{{A sequential particle filter method for static models}}.
\newblock \bibinfo{journal}{\emph{Biometrika}}  \bibinfo{volume}{89} (\bibinfo{date}{August} \bibinfo{year}{2002}), \bibinfo{pages}{539--552}.
\newblock
Issue 3.
\urldef\tempurl%
\url{https://doi.org/10.1093/biomet/89.3.539}
\showDOI{\tempurl}


\bibitem[Cusumano-Towner et~al\mbox{.}(2019)]%
        {PLDI:CSL19}
\bibfield{author}{\bibinfo{person}{Marco~F. Cusumano-Towner}, \bibinfo{person}{Feras~A. Saad}, \bibinfo{person}{Alexander~K. Lew}, {and} \bibinfo{person}{Vikash~K. Mansinghka}.} \bibinfo{year}{2019}\natexlab{}.
\newblock \showarticletitle{{Gen: A General-Purpose Probabilistic Programming System with Programmable Inference}}. In \bibinfo{booktitle}{\emph{Prog.\ Lang.\ Design and Impl.}} \emph{(\bibinfo{series}{PLDI'19})}.
\newblock
\urldef\tempurl%
\url{https://doi.org/10.1145/3314221.3314642}
\showDOI{\tempurl}


\bibitem[Del~Moral et~al\mbox{.}(2006)]%
        {JRSS:DDJ06}
\bibfield{author}{\bibinfo{person}{Pierre Del~Moral}, \bibinfo{person}{Arnaud Doucet}, {and} \bibinfo{person}{Ajay Jasra}.} \bibinfo{year}{2006}\natexlab{}.
\newblock \showarticletitle{{Sequential Monte Carlo Samplers}}.
\newblock \bibinfo{journal}{\emph{J.\ Royal Statistical Society}}  \bibinfo{volume}{68} (\bibinfo{date}{January} \bibinfo{year}{2006}).
\newblock
Issue 3.
\urldef\tempurl%
\url{https://www.jstor.org/stable/3879283}
\showURL{%
\tempurl}


\bibitem[Del~Moral et~al\mbox{.}(2007)]%
        {BS:DDJ07}
\bibfield{author}{\bibinfo{person}{Pierre Del~Moral}, \bibinfo{person}{Arnaud Doucet}, {and} \bibinfo{person}{Ajay Jasra}.} \bibinfo{year}{2007}\natexlab{}.
\newblock \showarticletitle{{Sequential Monte Carlo for Bayesian Computation}}.
\newblock \bibinfo{journal}{\emph{Bayesian Statistics}}  \bibinfo{volume}{8} (\bibinfo{year}{2007}).
\newblock


\bibitem[Del~Moral et~al\mbox{.}(2001)]%
        {RMTA:DKR01}
\bibfield{author}{\bibinfo{person}{Pierre Del~Moral}, \bibinfo{person}{L. Kallel}, {and} \bibinfo{person}{J. Rowe}.} \bibinfo{year}{2001}\natexlab{}.
\newblock \showarticletitle{{Modeling genetic algorithms with interacting particle systems}}.
\newblock \bibinfo{journal}{\emph{Revista De Matem{\'a}tica: Teor{\'i}a Y Aplicaciones}}  \bibinfo{volume}{8} (\bibinfo{year}{2001}), \bibinfo{pages}{19--77}.
\newblock
Issue 2.
\urldef\tempurl%
\url{https://doi.org/10.15517/rmta.v8i2.201}
\showDOI{\tempurl}


\bibitem[Dorigo et~al\mbox{.}(2008)]%
        {ANTS08}
\bibfield{editor}{\bibinfo{person}{Marco Dorigo}, \bibinfo{person}{Mauro Birattari}, \bibinfo{person}{Christian Blum}, \bibinfo{person}{Maurice Clerc}, \bibinfo{person}{Thomas St{\"u}tzle}, {and} \bibinfo{person}{Alan F.~T. Winfield}} (Eds.). \bibinfo{year}{2008}\natexlab{}.
\newblock \bibinfo{booktitle}{\emph{Ant Colony Optimization and Swarm Intelligence}}. \bibinfo{publisher}{Springer Berlin, Heidelberg}.
\newblock
\urldef\tempurl%
\url{https://doi.org/10.1007/978-3-540-87527-7}
\showDOI{\tempurl}


\bibitem[Drugan and Thierens(2003)]%
        {EA:DT03}
\bibfield{author}{\bibinfo{person}{M{\u a}d{\u a}lina~M. Drugan} {and} \bibinfo{person}{Dirk Thierens}.} \bibinfo{year}{2003}\natexlab{}.
\newblock \showarticletitle{{Evolutionary Markov Chain Monte Carlo}}. In \bibinfo{booktitle}{\emph{International Conference on Artificial Evolution}} \emph{(\bibinfo{series}{EA'03})}. \bibinfo{pages}{63--76}.
\newblock
\urldef\tempurl%
\url{https://doi.org/10.1007/978-3-540-24621-3_6}
\showDOI{\tempurl}


\bibitem[Dufays(2016)]%
        {Econometrics:Dufays16}
\bibfield{author}{\bibinfo{person}{Arnaud Dufays}.} \bibinfo{year}{2016}\natexlab{}.
\newblock \showarticletitle{{Evolutionary Sequential Monte Carlo Samplers for Change-Point Models}}.
\newblock \bibinfo{journal}{\emph{Econometrics}}  \bibinfo{volume}{4} (\bibinfo{date}{March} \bibinfo{year}{2016}).
\newblock
Issue 1.
\urldef\tempurl%
\url{https://doi.org/10.3390/econometrics4010012}
\showDOI{\tempurl}


\bibitem[Fr{\"u}hwirth-Schnatter(2004)]%
        {TEJ:FS04}
\bibfield{author}{\bibinfo{person}{Sylvia Fr{\"u}hwirth-Schnatter}.} \bibinfo{year}{2004}\natexlab{}.
\newblock \showarticletitle{{Estimating marginal likelihoods for mixture and Markov switching models using bridge sampling techniques}}.
\newblock \bibinfo{journal}{\emph{The Econometrics Journal}}  \bibinfo{volume}{7} (\bibinfo{date}{June} \bibinfo{year}{2004}), \bibinfo{pages}{143--167}.
\newblock
Issue 1.
\urldef\tempurl%
\url{https://doi.org/10.1111/j.1368-423X.2004.00125.x}
\showDOI{\tempurl}


\bibitem[Galeotti et~al\mbox{.}(2013)]%
        {ISSRE:GFA13}
\bibfield{author}{\bibinfo{person}{Juan~Pablo Galeotti}, \bibinfo{person}{Gordon Fraser}, {and} \bibinfo{person}{Andrea Arcuri}.} \bibinfo{year}{2013}\natexlab{}.
\newblock \showarticletitle{{Improving search-based test suite generation with dynamic symbolic execution}}. In \bibinfo{booktitle}{\emph{International Symposium on Software Reliability Engineering}} \emph{(\bibinfo{series}{ISSRE'13})}. \bibinfo{pages}{360--369}.
\newblock
\urldef\tempurl%
\url{https://doi.org/10.1109/ISSRE.2013.6698889}
\showDOI{\tempurl}


\bibitem[Gelman et~al\mbox{.}(2013)]%
        {book:GCS13}
\bibfield{author}{\bibinfo{person}{Andrew Gelman}, \bibinfo{person}{John~B. Carlin}, \bibinfo{person}{Hal~S. Stern}, \bibinfo{person}{David~B. Dunson}, \bibinfo{person}{Aki Vehtari}, {and} \bibinfo{person}{Donald~B. Rubin}.} \bibinfo{year}{2013}\natexlab{}.
\newblock \bibinfo{booktitle}{\emph{{Bayesian Data Analysis}}}.
\newblock \bibinfo{publisher}{Chapman and Hall/CRC}.
\newblock
\urldef\tempurl%
\url{https://doi.org/10.1201/b16018}
\showDOI{\tempurl}


\bibitem[Ghahramani(2015)]%
        {NATURE:Ghahramani15}
\bibfield{author}{\bibinfo{person}{Zoubin Ghahramani}.} \bibinfo{year}{2015}\natexlab{}.
\newblock \showarticletitle{{Probabilistic machine learning and artificial intelligence}}.
\newblock \bibinfo{journal}{\emph{Nature}}  \bibinfo{volume}{521} (\bibinfo{date}{May} \bibinfo{year}{2015}), \bibinfo{pages}{452--459}.
\newblock
\urldef\tempurl%
\url{https://doi.org/10.1038/nature14541}
\showDOI{\tempurl}


\bibitem[Gilks and Berzuini(2001)]%
        {JRSS:GB01}
\bibfield{author}{\bibinfo{person}{Walter~R. Gilks} {and} \bibinfo{person}{Carlo Berzuini}.} \bibinfo{year}{2001}\natexlab{}.
\newblock \showarticletitle{{Following a Moving Target-Monte Carlo Inference for Dynamic Bayesian Models}}.
\newblock \bibinfo{journal}{\emph{Journal of the Royal Statistical Society}}  \bibinfo{volume}{63} (\bibinfo{year}{2001}), \bibinfo{pages}{127--146}.
\newblock
Issue 1.


\bibitem[Goldberg(1989)]%
        {book:Goldberg89}
\bibfield{author}{\bibinfo{person}{David~E. Goldberg}.} \bibinfo{year}{1989}\natexlab{}.
\newblock \bibinfo{booktitle}{\emph{{Genetic Algorithms in Search, Optimization and Machine Learning}}}.
\newblock \bibinfo{publisher}{Addison-Wesley Longman Publishing Co., Inc.}
\newblock


\bibitem[Goodman et~al\mbox{.}(2008)]%
        {UAI:GMR08}
\bibfield{author}{\bibinfo{person}{Noah~D. Goodman}, \bibinfo{person}{Vikash~K. Mansinghka}, \bibinfo{person}{Daniel Roy}, \bibinfo{person}{Keith~A. Bonawitz}, {and} \bibinfo{person}{Joshua~B. Tenenbaum}.} \bibinfo{year}{2008}\natexlab{}.
\newblock \showarticletitle{{Church: A language for generative models}}. In \bibinfo{booktitle}{\emph{Uncertainty in Artificial Intelligence}} \emph{(\bibinfo{series}{UAI'08})}.
\newblock
\urldef\tempurl%
\url{https://dl.acm.org/doi/10.5555/3023476.3023503}
\showURL{%
\tempurl}


\bibitem[Goodman and Stuhlm{\"u}ller(2014)]%
        {misc:dippl}
\bibfield{author}{\bibinfo{person}{Noah~D. Goodman} {and} \bibinfo{person}{Andreas Stuhlm{\"u}ller}.} \bibinfo{year}{2014}\natexlab{}.
\newblock \bibinfo{title}{{The Design and Implementation of Probabilistic Programming Languages}}.
\newblock \bibinfo{howpublished}{Available on \url{http://dippl.org}}.
\newblock


\bibitem[Griffiths et~al\mbox{.}(2008)]%
        {kn:GKT08}
\bibfield{author}{\bibinfo{person}{Thomas~L. Griffiths}, \bibinfo{person}{Charles Kemp}, {and} \bibinfo{person}{Joshua~B. Tenenbaum}.} \bibinfo{year}{2008}\natexlab{}.
\newblock \showarticletitle{{Bayesian Models of Cognition}}.
\newblock In \bibinfo{booktitle}{\emph{The Cambridge Handbook of Computational Psychology}}. \bibinfo{publisher}{Cambridge University Press}.
\newblock
\urldef\tempurl%
\url{https://doi.org/10.1017/CBO9780511816772.006}
\showDOI{\tempurl}


\bibitem[Gu et~al\mbox{.}(2015)]%
        {NIPS:GGT15}
\bibfield{author}{\bibinfo{person}{Shixiang Gu}, \bibinfo{person}{Zoubin Ghahramani}, {and} \bibinfo{person}{Richard~E. Turner}.} \bibinfo{year}{2015}\natexlab{}.
\newblock \showarticletitle{{Neural Adaptive Sequential Monte Carlo}}. In \bibinfo{booktitle}{\emph{Neural Info.\ Processing Syst.}} \emph{(\bibinfo{series}{NIPS'15})}.
\newblock
\urldef\tempurl%
\url{https://dl.acm.org/doi/10.5555/2969442.2969533}
\showURL{%
\tempurl}


\bibitem[Hoffmann et~al\mbox{.}(2017)]%
        {POPL:HDW17}
\bibfield{author}{\bibinfo{person}{Jan Hoffmann}, \bibinfo{person}{Ankush Das}, {and} \bibinfo{person}{Shu-Chun Weng}.} \bibinfo{year}{2017}\natexlab{}.
\newblock \showarticletitle{{Towards Automatic Resource Bound Analysis for OCaml}}. In \bibinfo{booktitle}{\emph{Princ.\ of Prog.\ Lang.}} \emph{(\bibinfo{series}{POPL'17})}. \bibinfo{pages}{359--373}.
\newblock
\urldef\tempurl%
\url{https://doi.org/10.1145/3009837.3009842}
\showDOI{\tempurl}


\bibitem[Jasra et~al\mbox{.}(2007)]%
        {Biometrika:JSH07}
\bibfield{author}{\bibinfo{person}{Ajay Jasra}, \bibinfo{person}{David~A. Stephens}, {and} \bibinfo{person}{Christopher~C. Holmes}.} \bibinfo{year}{2007}\natexlab{}.
\newblock \showarticletitle{{Population-Based Reversible Jump Markov Chain Monte Carlo}}.
\newblock \bibinfo{journal}{\emph{Biometrika}}  \bibinfo{volume}{94} (\bibinfo{date}{December} \bibinfo{year}{2007}), \bibinfo{pages}{787--807}.
\newblock
Issue 4.


\bibitem[Kennedy and Eberhart(1995)]%
        {ICNN:KE95}
\bibfield{author}{\bibinfo{person}{James Kennedy} {and} \bibinfo{person}{Russell~C. Eberhart}.} \bibinfo{year}{1995}\natexlab{}.
\newblock \showarticletitle{{Particle swarm optimization}}. In \bibinfo{booktitle}{\emph{International Conference on Neural Networks}} \emph{(\bibinfo{series}{ICNN'95})}. \bibinfo{pages}{1942--1948}.
\newblock
\urldef\tempurl%
\url{https://doi.org/10.1109/ICNN.1995.488968}
\showDOI{\tempurl}


\bibitem[Kersten et~al\mbox{.}(2017)]%
        {CCS:KLP17}
\bibfield{author}{\bibinfo{person}{Rody Kersten}, \bibinfo{person}{Kasper Luckow}, {and} \bibinfo{person}{Corina~S. P{\u a}s{\u a}reanu}.} \bibinfo{year}{2017}\natexlab{}.
\newblock \showarticletitle{{POSTER: AFL-based Fuzzing for Java with Kelinci}}. In \bibinfo{booktitle}{\emph{Computer and Communications Security}} \emph{(\bibinfo{series}{CCS'17})}. \bibinfo{pages}{2511--2513}.
\newblock
\urldef\tempurl%
\url{https://doi.org/10.1145/3133956.3138820}
\showDOI{\tempurl}


\bibitem[Kozen(1981)]%
        {JCSS:Kozen81}
\bibfield{author}{\bibinfo{person}{Dexter Kozen}.} \bibinfo{year}{1981}\natexlab{}.
\newblock \showarticletitle{{Semantics of Probabilistic Programs}}.
\newblock \bibinfo{journal}{\emph{J.\ Comput.\ Syst.\ Sci.}}  \bibinfo{volume}{22} (\bibinfo{date}{June} \bibinfo{year}{1981}).
\newblock
Issue 3.
\urldef\tempurl%
\url{https://doi.org/10.1016/0022-0000(81)90036-2}
\showDOI{\tempurl}


\bibitem[Kumar and Kumar(2017)]%
        {AES:KK17}
\bibfield{author}{\bibinfo{person}{Vijay Kumar} {and} \bibinfo{person}{Dinesh Kumar}.} \bibinfo{year}{2017}\natexlab{}.
\newblock \showarticletitle{{An astrophysics-inspired Grey wolf algorithm for numerical optimization and its application to engineering design problems}}.
\newblock \bibinfo{journal}{\emph{Advances in Engineering Software}}  \bibinfo{volume}{112} (\bibinfo{date}{October} \bibinfo{year}{2017}), \bibinfo{pages}{231--254}.
\newblock
\urldef\tempurl%
\url{https://doi.org/10.1016/j.advengsoft.2017.05.008}
\showDOI{\tempurl}


\bibitem[Kwok et~al\mbox{.}(2005)]%
        {IROS:KFZ05}
\bibfield{author}{\bibinfo{person}{Ngaiming Kwok}, \bibinfo{person}{Gu Fang}, {and} \bibinfo{person}{Weizhen Zhou}.} \bibinfo{year}{2005}\natexlab{}.
\newblock \showarticletitle{{Evolutionary particle filter: re-sampling from the genetic algorithm perspective}}. In \bibinfo{booktitle}{\emph{International Conference on Intelligent Robots and Systems}} \emph{(\bibinfo{series}{IROS'05})}. \bibinfo{pages}{2935--2940}.
\newblock
\urldef\tempurl%
\url{https://doi.org/10.1109/IROS.2005.1545119}
\showDOI{\tempurl}


\bibitem[Lew et~al\mbox{.}(2023)]%
        {AISTATS:LMZ23}
\bibfield{author}{\bibinfo{person}{Alexander~K. Lew}, \bibinfo{person}{George Matheos}, \bibinfo{person}{Tan Zhi-Xuan}, \bibinfo{person}{Matin Ghavamizadeh}, \bibinfo{person}{Nishad Gothoskar}, \bibinfo{person}{Stuart Russell}, {and} \bibinfo{person}{Vikash~K. Mansinghka}.} \bibinfo{year}{2023}\natexlab{}.
\newblock \showarticletitle{{SMCP3: Sequential Monte Carlo with Probabilistic Program Proposals}}. In \bibinfo{booktitle}{\emph{Artificial Intelligence and Statistics}} \emph{(\bibinfo{series}{AISTATS'23})}. \bibinfo{pages}{7061--7088}.
\newblock


\bibitem[Liang and Wong(2000)]%
        {SS:LW00}
\bibfield{author}{\bibinfo{person}{Faming Liang} {and} \bibinfo{person}{Wing~Hung Wong}.} \bibinfo{year}{2000}\natexlab{}.
\newblock \showarticletitle{{Evolutionary Monte Carlo: Applications to $C_p$ Model Sampling and Change Point Problem}}.
\newblock \bibinfo{journal}{\emph{Statistica Sinica}}  \bibinfo{volume}{10} (\bibinfo{date}{April} \bibinfo{year}{2000}), \bibinfo{pages}{317--342}.
\newblock
Issue 2.


\bibitem[{LLVM Project}(2023)]%
        {misc:LibFuzzer}
\bibfield{author}{\bibinfo{person}{{LLVM Project}}.} \bibinfo{year}{2023}\natexlab{}.
\newblock \bibinfo{title}{{libFuzzer -- a library for coverage-guided fuzz testing}}.
\newblock \bibinfo{howpublished}{Available on \url{https://llvm.org/docs/LibFuzzer.html}}.
\newblock


\bibitem[Luckow et~al\mbox{.}(2017)]%
        {ICST:LKP17}
\bibfield{author}{\bibinfo{person}{Kasper Luckow}, \bibinfo{person}{Rody Kersten}, {and} \bibinfo{person}{Corina~S. P{\u a}s{\u a}reanu}.} \bibinfo{year}{2017}\natexlab{}.
\newblock \showarticletitle{{Symbolic Complexity Analysis Using Context-Preserving Histories}}. In \bibinfo{booktitle}{\emph{Int.\ Conf.\ on Softw.\ Testing, Verif.\ and Validation}} \emph{(\bibinfo{series}{ICST'17})}. \bibinfo{pages}{58--68}.
\newblock
\urldef\tempurl%
\url{https://doi.org/10.1109/ICST.2017.13}
\showDOI{\tempurl}


\bibitem[McNally et~al\mbox{.}(2012)]%
        {misc:MYG12}
\bibfield{author}{\bibinfo{person}{Richard McNally}, \bibinfo{person}{Ken Yiu}, \bibinfo{person}{Duncan Grove}, {and} \bibinfo{person}{Damien Gerhardy}.} \bibinfo{year}{2012}\natexlab{}.
\newblock \bibinfo{title}{{Fuzzing: The State of the Art}}.
\newblock \bibinfo{howpublished}{Available on \url{https://apps.dtic.mil/sti/citations/ADA558209}}.
\newblock


\bibitem[Noller et~al\mbox{.}(2018)]%
        {ISSTA:NKP18}
\bibfield{author}{\bibinfo{person}{Yannic Noller}, \bibinfo{person}{Rody Kersten}, {and} \bibinfo{person}{Corina~S. P{\u a}s{\u a}reanu}.} \bibinfo{year}{2018}\natexlab{}.
\newblock \showarticletitle{{Badger: Complexity Analysis with Fuzzing and Symbolic Execution}}. In \bibinfo{booktitle}{\emph{Int.\ Symp.\ on Softw.\ Testing and Analysis}} \emph{(\bibinfo{series}{ISSTA'18})}. \bibinfo{pages}{322--332}.
\newblock
\urldef\tempurl%
\url{https://doi.org/10.1145/3213846.3213868}
\showDOI{\tempurl}


\bibitem[Petsios et~al\mbox{.}(2017)]%
        {CCS:PZK17}
\bibfield{author}{\bibinfo{person}{Theofilos Petsios}, \bibinfo{person}{Jason Zhao}, \bibinfo{person}{Angelos~D. Keromytis}, {and} \bibinfo{person}{Suman Jana}.} \bibinfo{year}{2017}\natexlab{}.
\newblock \showarticletitle{{SlowFuzz: Automated Domain-Independent Detection of Algorithmic Complexity Vulnerabilities}}. In \bibinfo{booktitle}{\emph{Computer and Communications Security}} \emph{(\bibinfo{series}{CCS'17})}. \bibinfo{pages}{2155--2168}.
\newblock
\urldef\tempurl%
\url{https://doi.org/10.1145/3133956.3134073}
\showDOI{\tempurl}


\bibitem[Saad et~al\mbox{.}(2023)]%
        {ICML:SPH23}
\bibfield{author}{\bibinfo{person}{Feras~A. Saad}, \bibinfo{person}{Brian Patton}, \bibinfo{person}{Matthew~D. Hoffman}, \bibinfo{person}{Rif~A. Saurous}, {and} \bibinfo{person}{Vikash~K. Mansinghka}.} \bibinfo{year}{2023}\natexlab{}.
\newblock \showarticletitle{{Sequential Monte Carlo Learning for Time Series Structure Discovery}}. In \bibinfo{booktitle}{\emph{International Conference on Machine Learning}} \emph{(\bibinfo{series}{ICML'23})}. \bibinfo{pages}{29473--29489}.
\newblock


\bibitem[Thrun et~al\mbox{.}(2005)]%
        {book:ProbRobot05}
\bibfield{author}{\bibinfo{person}{Sebastian Thrun}, \bibinfo{person}{Wolfram Burgard}, {and} \bibinfo{person}{Dieter Fox}.} \bibinfo{year}{2005}\natexlab{}.
\newblock \bibinfo{booktitle}{\emph{{Probabilistic Robotics}}}.
\newblock \bibinfo{publisher}{MIT Press}.
\newblock
\urldef\tempurl%
\url{https://dl.acm.org/doi/10.5555/1121596}
\showURL{%
\tempurl}


\bibitem[Wang and Hoffmann(2019)]%
        {POPL:WH19}
\bibfield{author}{\bibinfo{person}{Di Wang} {and} \bibinfo{person}{Jan Hoffmann}.} \bibinfo{year}{2019}\natexlab{}.
\newblock \showarticletitle{{Type-guided worst-case input generation}}.
\newblock \bibinfo{journal}{\emph{Proceedings of the ACM on Programming Languages}} \bibinfo{volume}{3}, \bibinfo{number}{13} (\bibinfo{date}{January} \bibinfo{year}{2019}), \bibinfo{pages}{13:1--13:30}.
\newblock
Issue POPL.
\urldef\tempurl%
\url{https://doi.org/10.1145/3290326}
\showDOI{\tempurl}


\bibitem[Wigren et~al\mbox{.}(2018)]%
        {SYSID:WML18}
\bibfield{author}{\bibinfo{person}{Anna Wigren}, \bibinfo{person}{Lawrence~M. Murray}, {and} \bibinfo{person}{Fredrik Lindsten}.} \bibinfo{year}{2018}\natexlab{}.
\newblock \showarticletitle{{Improving the particle filter in high dimensions using conjugate artificial process noise}}. In \bibinfo{booktitle}{\emph{IFAC Symp.\ on System Identification}} \emph{(\bibinfo{series}{SYSID'18})}.
\newblock
\urldef\tempurl%
\url{https://doi.org/10.1016/j.ifacol.2018.09.207}
\showDOI{\tempurl}


\bibitem[Zalewski(2023)]%
        {misc:AFL}
\bibfield{author}{\bibinfo{person}{Michal Zalewski}.} \bibinfo{year}{2023}\natexlab{}.
\newblock \bibinfo{title}{{america fuzzy lop}}.
\newblock \bibinfo{howpublished}{Available on \url{https://lcamtuf.coredump.cx/afl/}}.
\newblock


\bibitem[Zhu et~al\mbox{.}(2018)]%
        {AWR:ZLM18}
\bibfield{author}{\bibinfo{person}{Gaofeng Zhu}, \bibinfo{person}{Xin Li}, \bibinfo{person}{Jinzhu Ma}, \bibinfo{person}{Yunquan Wang}, \bibinfo{person}{Shaomin Liu}, \bibinfo{person}{Chunlin Huang}, \bibinfo{person}{Kun Zhang}, {and} \bibinfo{person}{Xiaoli Hu}.} \bibinfo{year}{2018}\natexlab{}.
\newblock \showarticletitle{{A new moving strategy for the sequential Monte Carlo approach in optimizing the hydrological model parameters}}.
\newblock \bibinfo{journal}{\emph{Advances in Water Resources}}  \bibinfo{volume}{114} (\bibinfo{date}{April} \bibinfo{year}{2018}), \bibinfo{pages}{164--179}.
\newblock
\urldef\tempurl%
\url{https://doi.org/10.1016/j.advwatres.2018.02.007}
\showDOI{\tempurl}


\bibitem[Zhu et~al\mbox{.}(2022)]%
        {ACS:ZWC22}
\bibfield{author}{\bibinfo{person}{Xiaogang Zhu}, \bibinfo{person}{Sheng Wen}, \bibinfo{person}{Seyit Camtepe}, {and} \bibinfo{person}{Yang Xiang}.} \bibinfo{year}{2022}\natexlab{}.
\newblock \showarticletitle{{Fuzzing: A Survey for Roadmap}}.
\newblock \bibinfo{journal}{\emph{Comput. Surveys}} \bibinfo{volume}{54}, \bibinfo{number}{230} (\bibinfo{date}{September} \bibinfo{year}{2022}), \bibinfo{pages}{230:1--230:36}.
\newblock
Issue 11s.
\urldef\tempurl%
\url{https://doi.org/10.1145/3512345}
\showDOI{\tempurl}


\end{thebibliography}

%\clearpage
%\appendix
%\include{proofs}

\end{document}